%% file: main.tex
\documentclass[runningheads]{llncs}

\usepackage{eccv}

\usepackage{eccvabbrv}

\usepackage{wrapfig}
\usepackage{graphicx}
\usepackage{caption}
\usepackage{color}
\usepackage{multirow}
\usepackage{multicol}
\usepackage{float}
\usepackage{url}
\usepackage{array}
\usepackage{siunitx}
\usepackage{pifont}     
\usepackage{xcolor}
\usepackage{colortbl}
\usepackage{booktabs}

\newcommand{\red}[1]{{\color{red}#1}}

\newrobustcmd{\B}{\bfseries}                            
\newrobustcmd{\U}{\underline}
\newcommand{\xmark}{\ding{55}}%
\newcommand{\fade}{\textcolor[gray]{0.4}}
\definecolor{lavender}{rgb}{0.9, 0.9, 0.98}
\definecolor{lightgray}{rgb}{0.5, 0.5, 0.5}
\definecolor{beaublue}{rgb}{0.74, 0.83, 0.9}

\usepackage[accsupp]{axessibility}  

\usepackage{hyperref}

\usepackage{orcidlink}

\begin{document}

\title{PLOT: Pseudo-Labeling via Object Tracking\texorpdfstring{\\}{ }for Monocular 3D Object Detection} 

\titlerunning{PLOT}


\author{Seokyeong Lee\thanks{Equal contribution (Work done in KIST).}\inst{2}\orcidlink{0000-0003-3705-3723} \and
Sithu Aung\inst{\star3}\orcidlink{0009-0006-4151-8545} \and
Junyong Choi\inst{4}\orcidlink{0009-0005-8402-9365} \and
\\
Seungryong Kim\inst{2}\orcidlink{0000-0003-2927-6273} \and
Ig-Jae Kim\inst{1,5}\orcidlink{0000-0002-2741-7047} \and
Junghyun Cho\inst{1,5,6}\orcidlink{0000-0003-1913-8037}}

\authorrunning{S.~Lee et al.}

\institute{
\leavevmode\inst{1} Korea Institute of Science and Technology (KIST) \quad
\leavevmode\inst{2} KAIST AI \\
\leavevmode\inst{3} VRG, FEE, Czech Technical University in Prague \quad
\leavevmode\inst{4} Samsung Research \\
\leavevmode\inst{5} AI-Robotics, KIST School, University of Science and Technology (UST) \\
\leavevmode\inst{6} Yonsei-KIST Convergence Research Institute, Yonsei University
\email{
\{shapin94,seungryong\}@kaist.ac.kr}\
\email{sithu.aung@fel.cvut.cz}\
\email{\{drjay,jhcho\}@kist.re.kr}
}

\maketitle
\input{sec/0_abstract}
\input{sec/1_introduction}

\input{sec/2_related_works}
\input{sec/3_method}
\input{sec/4_experiments}
\input{sec/5_conclusion}

\section*{Acknowledgements}
This work was partly supported by Institute of Information \& communications Technology Planning \& Evaluation (IITP) grant funded by the Korea government(MSIT)(RS-2023-00227592, Development of 3D Object Identification Technology Robust to Viewpoint Changes), the Korea Institute of Police Technology (KIPoT) funded by the Korean National Police Agency \& Ministry of the Interior and Safety (RS-2024-00405100), and the Korea Institute of Science and Technology (KIST) Institutional Program (Project No. 62E0062).
This work was also partly funded by the
Czech Science Foundation (GACR) JUNIOR STAR Grant
No. 22-23183M (supporting S.Aung).

%
%


\newpage
\clearpage
\appendix
\begin{center}
    \begingroup
    \centering
    {\LARGE \bf Supplementary Material \par}
    \endgroup
\end{center}

\setcounter{section}{0}
\setcounter{equation}{0}
\setcounter{figure}{0}
\setcounter{table}{0}

\renewcommand{\thesection}{A\arabic{section}}
\renewcommand{\theequation}{A\arabic{equation}}
\renewcommand{\thefigure}{A\arabic{figure}}
\renewcommand{\thetable}{A\arabic{table}}

\input{sec/6_appendix}

\newpage
\clearpage
\bibliographystyle{splncs04}
\bibliography{main}

\end{document}

%% file: sec/0_abstract.tex
\begin{abstract}
Monocular 3D object detection is crucial for scalable perception across fields like autonomous driving, robotics, and surveillance. However, progress is hindered by limited 3D annotations and the inherent ambiguity of single-image geometry.
Existing methods often rely on strong geometric assumptions or carefully curated datasets, which limit their applicability to real-world scenarios.
In this paper, we present \textbf{PLOT} (\textbf{P}seudo-\textbf{L}abeling via \textbf{O}bject \textbf{T}racking), a framework that generates 3D annotations from monocular videos without auxiliary sensors or model retraining. 
PLOT tracks object and background trajectories to estimate camera motion and perform object association in pose-unknown settings. 
These trajectories provide point correspondences that align frame-wise pseudo-LiDARs, which are then fused via simple optimization into a unified object shape robust to occlusion and viewpoint shifts.
Recognizing temporal coherence as a fundamental requirement for reliable shape fusion and video perception, we design a global object memory that preserves consistent object identities across frames. 
PLOT achieves robust annotation quality and strong generalization on both M3OD video benchmarks and in-the-wild videos, proving its effectiveness across diverse and unconstrained domains.
Project page: \url{https://plot-eccv.github.io}.
  \keywords{Pseudo-labeling, Monocular 3D Object Detection}
\end{abstract}

%% file: sec/1_introduction.tex
\begin{figure*}[t]
    \centering
    \captionsetup{font={small}}
    \includegraphics[width=1.\textwidth]{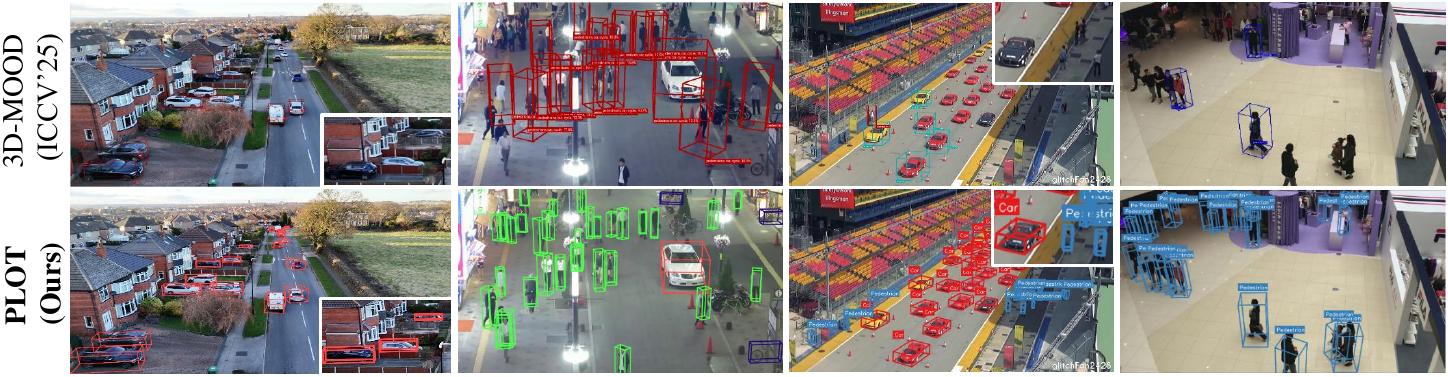}
    \caption{Zero-shot estimation on MOT17~\cite{MOT16} and Pexels~\cite{pexels2025}: Predictions from (top) an open-set M3OD model~\cite{yang20253d} and (bottom) our pseudo-labeling method. 
    Our method performs well on out-of-domain camera views (zoom in for a better view).}
    \label{fig:cubercnn_results}
\end{figure*}

\section{Introduction}
\label{sec:introduction}

Monocular 3D object detection (M3OD) aims to recover the 3D geometry, pose, and extent of objects from a single RGB camera, offering a cost-effective alternative to LiDAR-based perception systems~\cite{shi2019pointrcnn, Hu_2022_CVPR, pan2024pretraining}. 
Despite substantial progress, M3OD remains fundamentally ill-posed due to depth--scale ambiguity and the lack of direct geometric supervision, which has led most methods to rely on curated benchmarks collected in sensor-rich environments, such as autonomous driving~\cite{kitti, waymo, nuscenes} and indoor scenes~\cite{dai2017scannet, baruch2021arkitscenes}. 
Consequently, models developed under these settings often exhibit limited generalization when deployed in unconstrained scenarios with moving cameras, unknown poses, and diverse viewpoints, as evidenced by their poor zero-shot performance on in-the-wild videos~\cite{MOT16, pexels2025} (Fig.~\ref{fig:cubercnn_results}).

To alleviate the scarcity of high-quality 3D annotations, pseudo-labeling and weakly supervised approaches have been proposed to generate 3D supervision without explicit LiDAR or multi-view data~\cite{dpl, huangtraining, liu2024vsrd}. 
However, most existing pipelines operate on single images, limiting their ability to resolve occlusions and disambiguate object scale and orientation. 
These single-frame formulations produce partial and noisy geometric observations, which propagate systematic errors into downstream training and evaluation. 
Moreover, many methods rely on sensor poses or hand-crafted geometric heuristics~\cite{skvrna2025monosowa, huangtraining}, restricting their scalability and robustness across domains.

In this paper, we propose \textbf{PLOT} (\textbf{P}seudo-\textbf{L}abeling via \textbf{O}bject \textbf{T}racking), a framework that \textbf{generates reliable 3D annotations directly from monocular videos without requiring auxiliary sensors, camera poses, or model retraining. }
Monocular videos are abundant and encode rich geometric cues over time, offering a natural source of supervision beyond single images. 
However, effectively leveraging such videos introduces two fundamental challenges that have remained underexplored in prior work.

The first challenge is maintaining long-term object identity consistency in pose-unknown videos. In realistic settings, clutter, occlusion, and intermittent detection failures frequently lead to fragmented trajectories and identity switches, degrading the quality of aggregated labels. 
PLOT addresses this challenge by aligning frame-wise object observations through dense point tracking~\cite{harley2025alltracker} and introducing a \emph{Global Object Memory} (GOM) that models object persistence over time, enabling robust association even under occlusion and detector noise.

The second challenge concerns shape incompleteness and its impact on estimating key 3D attributes such as object size, orientation, and position. 
While recent work~\cite{skvrna2025monosowa} attempts to leverage temporal aggregation from monocular videos, they often rely on simplified matching strategies or limited association mechanisms, resulting in incomplete and fragmented object geometry. 
Consequently, downstream geometric estimators operate on partial observations that do not adequately capture the full object extent. 
For instance, orientation estimation methods such as PCA-based inference implicitly assume sufficiently complete geometry; when applied to sparse or occluded point sets, these assumptions break down, leading to biased or unstable predictions. 
PLOT addresses this issue by aggregating object observations across time via tracking-based motion estimation and trajectory-guided shape fusion, progressively recovering more complete object geometry. Unlike global scene reconstruction methods~\cite{leroy2024mast3r, wang2025vggt, zhang2024monst3r, wang2025cut3r}, which primarily target static backgrounds and lack object-level reasoning, our approach enables object-centric shape completion tailored to dynamic instances.

\begin{figure*}[t]  
\begin{center}
\centering
    \includegraphics[width=1.0\textwidth]{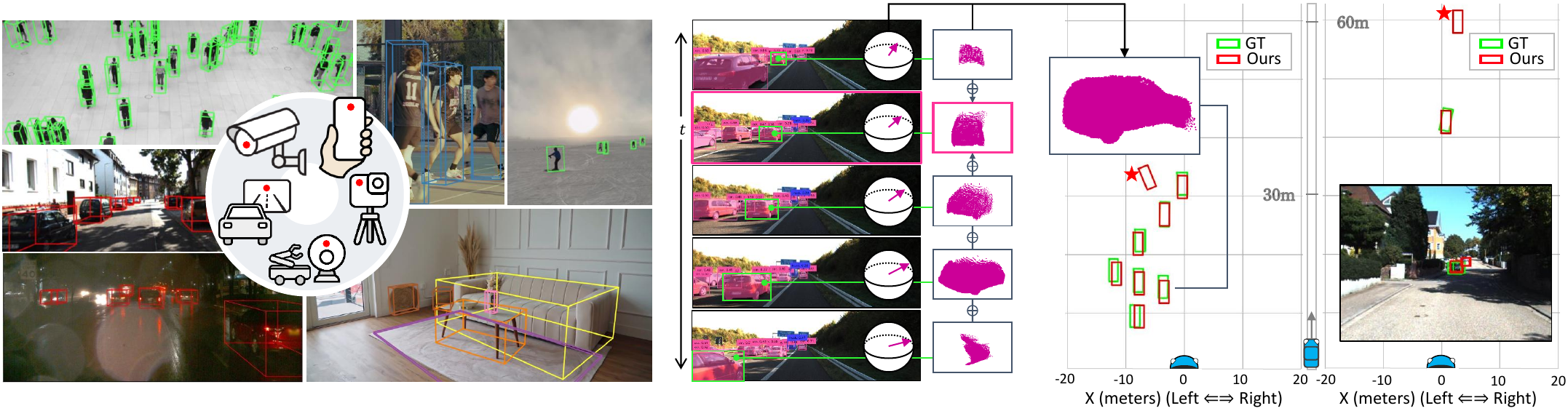}
    \begin{tabular}{p{0.38\textwidth} p{0.6\textwidth}}
            \scriptsize (a) Labeling results in diverse domains & \scriptsize (b) Multi-frame aggregation \& BEV comparison with GT
    \end{tabular}   
    \caption{PLOT generates accurate 3D labels directly from monocular videos without requiring auxiliary sensors or training, as illustrated in (a) qualitative results across diverse scenarios. Furthermore, (b) our object tracking and aggregation pipeline produces shape-complete pseudo-LiDARs, yielding BEV maps comparable to ground truth and can identify miss-labeled objects (marked with a red star).}
    \label{fig:teaser} 
    \end{center}
\end{figure*}

We validate PLOT on multiple M3OD benchmarks~\cite{kitti, kitti360, waymo} and unconfined monocular videos~\cite{MOT16, ding2023mose, pexels2025}, demonstrating strong performance across diverse domains and camera motions (see Fig.~\ref{fig:teaser}). 
Overall, PLOT provides a practical and scalable solution for generating high-quality 3D supervision in pose-unknown, real-world video settings, expanding the applicability of M3OD beyond curated sensor environments.

%% file: sec/2_related_works.tex
\section{Related Work}
\label{sec:rel_work}
\subsubsection{Supervised Monocular 3D Object Detection~(M3OD).}
Early supervised methods~\cite{brazil2019m3d, wang2021tent, peng2022didm3d, gupnet, jia2023monouni} achieve progress in curated settings but remain restricted by scarce 3D annotations. To alleviate this, several approaches use external data such as video~\cite{brazil2020kinematic, wang2022dfm}, LiDAR~\cite{huang2022monodtr}, or CAD models~\cite{liu2021autoshape, Parihar_2025_CVPR}.  
Recent works move toward open-set detection, exemplified by CubeR-CNN~\cite{brazil2023omni3d} and successors~\cite{yao2024open, zhang2025detany3d, yang20253d}, exploiting large-scale benchmarks~\cite{brazil2023omni3d} and 2D foundation models~\cite{ravi2024sam, oquab2023dinov2}. While these broaden domain coverage, they remain tied to 3D supervision and struggle to generalize consistently in unconstrained videos (Fig.~\ref{fig:cubercnn_results}).


\subsubsection{Weakly-supervised M3OD.}
Given the challenges of cost and scarcity of 3D annotations, several works~\cite{VS3D, qin2021monogrnet, han2024weakly, jiang2024weaklysingle} have pioneered weakly-supervised M3OD.
WeakM3D~\cite{peng2022weakm3d} estimates 3D attributes given raw LiDAR points, while WeakMono3D~\cite{weakmono3d} and subsequent work~\cite{han2024weakly} utilize multi-view constraints.
SKD-WM3D~\cite{jiang2024weaklysingle} proposes a self-teaching pipeline to lift 2D features into 3D space using depth completion, while VGW-3D~\cite{huang2024vgw3d} uses perspective-invariant error prediction and decoupled visual guidance.
Importantly, MonoGRNet~\cite{qin2021monogrnet} introduces a general-purpose model that utilizes video priors.
However, these methods require sophisticated priors, which limit their scalability. GGA~\cite{zhang2024gga} addresses this issue by introducing a generalizable weakly-supervised M3OD framework that leverages general geometric priors and 2D ground-truths. VSRD~\cite{liu2024vsrd} proves the efficacy of pseudo-labeling and silhouette rendering in weakly-supervised M3OD.


\subsubsection{Tracking for 3D Perception.}
Tracking has recently gained attention in video-based 3D perception for its potential to model scene dynamics over time.
Seurat~\cite{Seurat} and ViPE~\cite{huang2025vipe} utilize dense tracking across frames to enhance 3D perception in dynamic scenes. While Seurat focuses on depth map estimation, ViPE integrates tracking signals for broader scene-level understanding. However, neither method addresses object-level geometry or pose.
Vid2CAD~\cite{maninis2022vid2cad} targets CAD model alignment using multi-frame tracking, but focuses on aligning indoor objects to canonical CAD poses without addressing real-world clutter.
In contrast, our approach uses off-the-shelf dense tracker~\cite{harley2025alltracker} to associate object observations over time and estimate relative camera motion—enabling pose-free, training-free 3D labeling from monocular videos without bundle adjustment or calibration.

\subsubsection{Pseudo-labeling for M3OD.}
To address the scarcity of 3D annotations, pseudo-labeling has emerged as a viable alternative for enabling M3OD without explicit 3D supervision. In particular, VSRD~\cite{liu2024vsrd} introduced a multi-view-based auto-labeling approach, showing that reliable pseudo-labels can enhance weakly supervised training. 
OVM3D-Det~\cite{huangtraining}, meanwhile, proposes a zero-shot labeling framework that utilizes open-vocabulary 2D detectors~\cite{metric3dv2, ren2024grounded} and LLM priors~\cite{achiam2023gpt} to estimate 3D box attributes without training. However, it operates on single images and thus remain sensitive to occlusions, noise, and scale ambiguities, limiting their applicability in more complex or dynamic scenes. 
MonoSOWA~\cite{skvrna2025monosowa}, a recent work, similarly leverages per-frame 2D mask association for pseudo-label generation, but the reliance on known camera poses with naive mask-based tracking limits its scalability, and does not address cluttered environments explicitly.

%% file: sec/3_method.tex
\section{Method}
\label{sec:methods}
\begin{figure*}[!t]
    \centering
    \includegraphics[width=\linewidth]{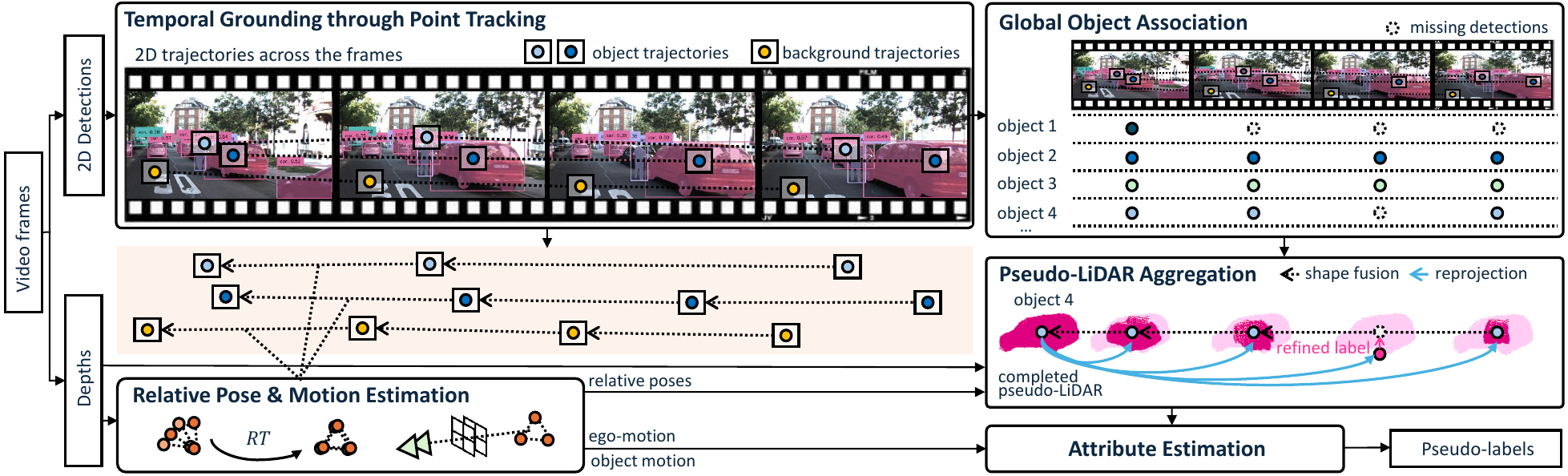}  
    \caption{Overall architecture of PLOT. Given monocular videos, we extract 2D detections and depth, and track points to obtain temporally grounded correspondences. These are used to estimate relative poses and camera motions across frames, enabling shape fusion and orientation estimation. A global object association module refines trajectories and recovers missing instances. Finally, completed pseudo-LiDARs are reprojected to each frame for consistent 3D attribute annotation.}
    \label{fig:framework}
\end{figure*}

We propose PLOT, a framework for generating 3D annotations from monocular videos without auxiliary sensors or model retraining.  
As illustrated in Fig.~\ref{fig:framework}, PLOT leverages off-the-shelf detectors and depth estimators to extract 2D masks and metric depth maps, which are combined via dense point tracking to form temporally grounded correspondences (Sec.~\ref{subsec:tracking}).  
These are used to estimate relative poses through point-based registration (Sec.~\ref{subsec:relative_pose}), enabling both shape fusion and motion analysis.  
To maintain identity consistency and recover missed instances, we introduce a global object memory (GOM) that refines labels over time (Sec.~\ref{subsec:goam}).  
Finally, object pseudo-LiDARs are constructed by aggregating registered points across frames and projecting them back to each time frame for consistent attribute estimation (Sec.~\ref{subsec:shapefusion}).

\subsection{Temporal Grounding through Point Tracking}
\label{subsec:tracking}

To enable robust object association and temporally grounded geometric reasoning, we distinguish between two types of 2D object mask instances. \textit{Predicted masks} are obtained at each frame via an open-vocabulary detector~(GSAM)~\cite{ren2024grounded}, while \textit{tracked masks} are generated by propagating previous-frame instances using a point-based 2D tracker~\cite{harley2025alltracker}. The former reflects per-frame shape variations, whereas the latter preserves point-level correspondences necessary for relative pose estimation and shape fusion.

Specifically, given the $k$-th object mask $M_k^t$ from an open-vocabulary 2D detector~\cite{ren2024grounded} in each frame image $I_t$ where $t \in \mathcal{T}$ (the full set of video time steps), we generate tracked masks $M_k^{\acute{t} \leftarrow t}$ at other time steps $\acute{t}$ by tracking $M_k^t$ to $\acute{t}$ using a dense point tracker~\cite{harley2025alltracker}:
\begin{equation}
    M_k^{\acute{t} \leftarrow t} = T_{\acute{t} \leftarrow t}(M_k^t), \quad \acute{t} \in \mathcal{T} \setminus \{t\}, \quad k \in \mathcal{K}^t.
\end{equation}
Here, $\mathcal{K}^t$ represents the total number of objects predicted at time $t$, and $T_{{\acute{t}} \leftarrow t}(\cdot)$ denotes dense tracking from $t$ to $\acute{t}$. Note that $\mathcal{K}^t$ can vary between frames due to missing labels caused by distance- or occlusion-related challenges, or simply detector errors.

Thus, the predicted and tracked masks may diverge, despite originating from the same stem.
We therefore perform explicit matching to associate them and maintain consistent identity over time. 
Using frames from two distinct time steps $s$ and $t$, we apply bipartite matching using the Hungarian algorithm~\cite{kuhn1955hungarian} between the set of tracked masks and the predicted masks, as follows:
\begin{equation}
    \hat{\sigma} = \arg\max_{\sigma \in \mathcal{S}_n} \sum_{i} \text{IoU}(M^{s \leftarrow t}_i, M^{s}_{\sigma(i)}),
\label{eq:hungarianmatching} 
\end{equation}
where $\hat{\sigma}$ represents the optimal assignment of predicted masks to tracked masks, determined by maximizing the sum of the Intersection of Union (IoU) scores, while $\mathcal{S}_n$ denotes the set of all possible permutations of assignments. 
Tracked masks are used to estimate rigid motion, while matched predicted masks serve as the basis for shape fusion.

In parallel, we track background points to infer camera motion and disentangle objects from static scene structure. Specifically, we sample background masks $M^t_{bg}$ from regions outside object masks and track them over time as follows:
\begin{equation}
    M^{\acute{t} \leftarrow t}_{bg} = T_{\acute{t} \leftarrow t}(M^t_{bg}),  \quad \acute{t} \in \mathcal{T} \setminus \{t\}, \quad M_{bg}^t \subset I_t \setminus M^t, \quad M^t = \bigcup_{k \in \mathcal{K}^t} M_{k}^t.
\end{equation}
These object and background correspondences are used to determine the motion and dynamic states of both the camera and objects, as demonstrated below.

\subsection{Pose and Motion Estimation via Point Trajectories}
\label{subsec:relative_pose}
In addition to object shape fusion, which relies on relative pose estimation across frames, it is equally important to recover the motion states of the camera. Accurate camera motion estimation enables downstream tasks such as object orientation reasoning, trajectory prediction, and relative positioning.
To achieve this, we derive the trajectories of both background points and object points across consecutive frames. 
Background point trajectories provide constraints for robust ego-motion estimation, while object point trajectories additionally capture individual object dynamics.

\subsubsection{Camera Motion Estimation via Background Trajectories.}
\label{subsec:cameramotion}
To estimate the camera motion, we leverage the trajectories of the background points, which are lifted to 3D using monocular depth predictions~\cite{unidepth}.
This yields a set of 3D motion trajectories $\{ \mathbf{p}_i^t | i \in M_{bg}^t, t \in \mathcal{T} \}$.
The camera motion between a reference frame $r$ (typically the first frame) and each frame $t$ is parameterized by rotation $\mathbf{R}_c^t \in SO(3)$ and translation $\mathbf{t}_c^t \in \mathbb{R}^3$.
We recover this motion via Procrustes alignment \cite{luo1999procrustes}:
\begin{equation}
    \operatorname*{arg\,min}_{s_c^t\mathbf{R}_c^t,\mathbf{t}_c^t}\; \sum_{i \in M_{bg}^t} \mathcal{M} \: \mathcal{V} \: || \mathbf{p}_i^r - (s_c^t\mathbf{R}_c^t \mathbf{p}_i^{r \leftarrow t} + \mathbf{t}_c^t)||^2,
\label{eq:cameramotion}
\end{equation}
where $\mathcal{M}$ is a binary mask that suppresses background points with unreliable depth estimates (e.g., beyond \SI{50}{\metre}), and $\mathcal{V}$ indicates mutual visibility of point $i$ in both frames $t$ and $r$.
The scale term $s_c^t$ is optional: it can be activated only when monocular depth predictions exhibit noticeable temporal scale drift or flicker; otherwise, we fix $s_c^t = 1$ and recover a purely rigid transformation.
This flexibility allows PLOT to remain robust across different depth estimators while avoiding unnecessary degrees of freedom when depth is already stable.
The estimated camera motion provides a temporally coherent reference frame, which is later used to transform object poses by compensating for ego-motion.

\subsubsection{Object Pose and Motion Estimation via Object Trajectories.}
Similarly to camera motion, we estimate object motion by computing frame-to-frame registration using reliable point correspondences within each object mask $M_k^t$ and its tracked counterpart $M_k^{r \leftarrow t}$:
\begin{equation}
    \operatorname*{arg\,min}_{s^t\mathbf{R}^t,\mathbf{t}^t}\; \sum_{i \in M_k^t} \mathcal{V} \: || \mathbf{p}_i^r - (s^t\mathbf{R}^t \mathbf{p}_i^{r \leftarrow t} + \mathbf{t}^t)||^2,
\label{eq:registration}
\end{equation}
where $\mathbf{R}^t$, $\mathbf{t}^t$ and $s^t$ denote the object's rotation, translation, and scale (optional).
The resulting rigid or similarity transformation is later used for temporal aggregation and shape fusion Sec.~\ref{subsec:shapefusion}.

Using the camera motion from Eq.~\ref{eq:cameramotion}, each object's local trajectory is transformed into the reference frame's coordinate to obtain a world-aligned trajectory:
\begin{equation}
    \hat{\mathbf{p}}_i^t = s_c^t\mathbf{R}_c^t\mathbf{p}_i^t + \mathbf{t}_c^t.
\end{equation}
The motion status of an object is then determined by its displacement between adjacent frames, i.e., $|| \hat{\mathbf{p}}_i^s - \hat{\mathbf{p}}_i^t ||$. 
If the displacement is negligible, the object is treated as static; otherwise, it is considered dynamic.
For dynamic objects, orientation is inferred from the direction of motion in world space.
Since driving scenes (e.g., KITTI~\cite{kitti}) predominantly exhibit motion along the ground plane, we compute the azimuth angle from trajectory changes:
\begin{equation}
    \theta_{k,\text{dyna}}^t = \mathrm{arctan2} \left( \frac{\hat{\mathbf{p}}_i^s(z) - \hat{\mathbf{p}}_i^t(z)}{\hat{\mathbf{p}}_i^s(x) - \hat{\mathbf{p}}_i^t(x)} \right).
\end{equation}
For static objects, we estimate orientation using principal component analysis (PCA) on the fused pseudo-LiDAR from Sec.~\ref{subsec:shapefusion}, assuming the major axis (highest variance) aligns with the object’s longitudinal axis. 
Unlike prior work~\cite{huangtraining}, our method benefits from denser pseudo-LiDAR from multiple observations, enabling PCA to yield more precise orientation estimates.
Per-frame orientation estimates are derived by correcting orientation based on ego-motion as follows:
\begin{equation}
    \theta_{k, \text{static}}^{t} = \Phi^{-1}\Big((\mathbf{R}_c^t)^{T}\Phi(\theta_k^{r})\Big),
\end{equation}
where $\Phi$ denotes the mapping from a yaw angle to a rotation matrix, and $\Phi^{-1}$ is its inverse.

\begin{wrapfigure}[13]{r}{0.5\textwidth}
    \vspace{-8mm}
    \centering
    \captionsetup{font={small}, skip=8pt}
    \includegraphics[width=0.48\textwidth]{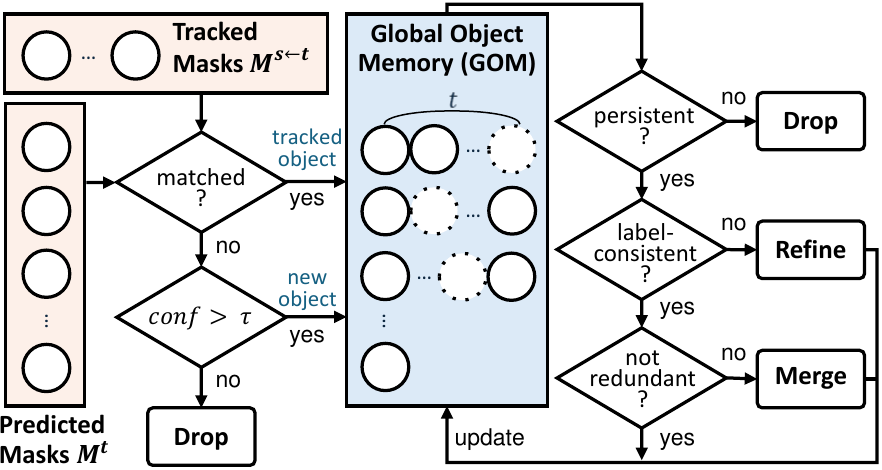}
    \caption{Global object association pipeline. Our memory-based object association consolidates noisy frame-level predictions into globally consistent tracks.}
    \label{fig:goam}

\end{wrapfigure}

\subsection{Global Object Memory}
\label{subsec:goam}


Although 2D point trackers~\cite{harley2025alltracker, karaev2025cotracker} and detectors~\cite{ren2024grounded} are effective per frame, occlusions and clutter still cause missed detections and visibility gaps, leading to fragmented trajectories. 
These issues manifest as identity switches and noisy pseudo-LiDARs from incorrect point aggregation.
To address this, we analyze common failure modes and introduce a Global Object Memory (GOM) that maintains object hypotheses and enforces consistent associations across frames.

\subsubsection{Failure Modes in Video Object Tracking.}
To examine the limitations of naive tracking-based video understanding, we analyze two dominant failure modes in real-world videos.
First, detection dropouts and label noise arise under occlusion, distance, or challenging conditions, where detectors may drop masks or assign inconsistent labels to the same object over time.
Second, identity switches in cluttered scenes occur when multiple objects enter, exit, or overlap within a frame, confusing local matching~(Eq.~\ref{eq:hungarianmatching}). 
In addition, temporary occlusions often reduce the visibility of tracked points below the tracker’s confidence threshold, causing the pipeline to prematurely terminate the trajectory-even though the object remains present-resulting in fragmented tracks.
Under these conditions, naive mask-level associations tend to assign new identities to previously observed objects or, conversely, merge distinct objects into a single track.
Such failure cases highlight the limits of local heuristics and motivate global, temporally structured association; qualitative examples and further analysis are provided in the supplementary material.

\subsubsection{Label Refinement via Global Object Memory.}
We introduce a Global Object Memory (GOM) that consolidates frame-level predictions into coherent object tracks across frames. As illustrated in Fig.~\ref{fig:goam}, GOM maintains persistent entries, each representing a unique object observed over time. Predicted masks $M^t$ are matched to memory entries based on spatial overlap with the tracked masks $M^{s\leftarrow t}$; if no match is found, a new entry is provisionally created only if the prediction confidence is greater than the detection threshold $\tau$.

GOM updates object entries by checking 1) whether the object persists across frames through point tracking, 2) whether the class labels and geometric attributes remain consistent, and 3) whether multiple memory entries correspond to the same object. Based on these checks, GOM either 1) discards unstable or incoherent instances, 2) refines entries by updating their attributes with dominant evidence, or 3) merges redundant entries. This refinement enforces long-term identity consistency and improves the reliability of object-level 3D annotations. 

In summary, GOM addresses two complementary sources of discontinuity: 1) breaks in temporal continuity caused by missed or duplicated masks from the detector, and 2) fragmented trajectory segments that arise when occlusion reduces point visibility and interrupts tracker propagation.



\subsection{Trajectory-guided Object Shape Fusion}
\label{subsec:shapefusion}
Using the globally associated object tracks constructed in Sec.~\ref{subsec:goam}, we build complete pseudo-LiDARs by aggregating object observations over time. 
Rather than requiring absolute camera poses, we perform shape fusion using relative poses derived from point trajectories in Sec.~\ref{subsec:relative_pose}.

For each object, we first identify the reference frame image $I_{r}$ that minimizes the total registration error between matched frames, using Eq.~\ref{eq:registration}. We then align all matched object masks to this target frame using their estimated relative rotation and translation:
\begin{equation}
    \hat{\mathbf{P}}_k = \bigcup_{t \in \mathcal{T}} s^{t \rightarrow r}\mathbf{R}^{t \rightarrow r}\mathbf{P}_k^t + \mathbf{t}^{t \rightarrow r},
\label{eq:pseudolidar}
\end{equation}
where $\mathbf{P}_k^t$ denotes the 3D points extracted from the $k$-th object mask at time $t$, and $t \rightarrow r$ indicates the transformation from $t$ to the reference frame $r$.
The resulting point cloud $\hat{\mathbf{P}}_k$ serves as a completed pseudo-LiDAR representation, fusing multiple partial views into a consistent shape.

To ensure reliable 3D annotations across all frames, we further project the completed pseudo-LiDAR $\hat{\mathbf{P}}_k$ back into each frame’s coordinate system using the inverse of the same relative transformation.

%% file: sec/4_experiments.tex
\begin{table*}[t]
    \caption{
        Detection results for the `Car' class on the KITTI~\cite{kitti} validation set. `Attr-Net': an additional attribute estimation network. `Depth': ground-truth depth map used in training. `Depth${\dagger}$': KITTI-trained depth completion model. `VPS': Video, Pose and Shape prior. Both \fade{supervised and open-set} M3OD models use 3D ground-truth data of KITTI in their training. 
    }
    \centering
    \resizebox{\linewidth}{!}
    {\input{tab/tab1_kitti}}
    \label{tab:kitti} 
\end{table*}

\section{Experiments}
In this section, we evaluate PLOT on standard monocular 3D object detection (M3OD) video benchmarks—KITTI~\cite{kitti}, KITTI-360~\cite{kitti360}, and Waymo~\cite{waymo}—and compare it against recent open-set detectors, weakly-supervised and pseudo-labeling methods that rely on driving-specific priors or pose estimates. While our labeler is designed for open-world settings, existing benchmarks with video inputs are constrained to driving scenes, limiting the scope of evaluation.
Thus, we provide qualitative comparisons on in-the-wild videos~\cite{ding2023mose, MOT16, pexels2025} to assess the generalization of PLOT beyond vehicle-centric environments. 
Finally, we present ablations and runtime comparisons to validate each component.
Further details about experimental setup and evaluation metrics are reported in the supplementary material.

\begin{table*}[t]
    \caption{
         Detection results for the `Car' class on the KITTI-360~\cite{kitti360} test set.
    }
    \centering
    \resizebox{0.95\linewidth}{!}
    {\input{tab/tab2_kitti360}}
    \label{tab:kitti360} 
\end{table*}


\subsection{Experiments on Driving Benchmarks}
In this subsection, we compare our method with various M3OD works (listed on the left of each table) on multiple benchmark datasets. The methods are grouped as \textbf{Sup.} (fully-supervised), \textbf{Weak.} (weakly-supervised), \textbf{Open.} (open-set), and \textbf{Pseudo.} (pseudo-labeling). For pseudo-labeling methods, including ours, results are obtained by training MonoDETR~\cite{zhang2023monodetr} on the generated labels with the same default configuration across all methods to ensure fairness; additional details are provided in the supplementary material. As most M3OD methods focus on the `Car' and `Pedestrian' classes, our main paper reports quantitative results primarily for these categories. 
Additional qualitative results and further analyses are provided in the supplementary material.

\begin{table*}[t]
    \caption{
         Detection results for the `Pedestrian' class on KITTI~\cite{kitti} validation set.
    }
    \centering
    \resizebox{\linewidth}{!}
    {\input{tab/atab1_kitti_ped}}
    \label{tab:kitti_ped} 
\end{table*}

\begin{table*}[t]
    \caption{
     Detection results for the `Vehicle' class on the Waymo~\cite{waymo} validation set.
    }
    \centering
    \resizebox{\linewidth}{!}
    {\input{tab/tab3_waymo}}
    \label{tab:waymo} 
\end{table*}

\subsubsection{KITTI \& KITTI-360 (Car).}  
Tab.~\ref{tab:kitti} and Tab.~\ref{tab:kitti360} provide quantitative results on the KITTI and KITTI-360 datasets for the `Car' class. For clarity, the label column lists both 2D ground-truth annotations (GT-2D) and predictions from 2D detectors~\cite{mvit2, ren2024grounded}, used in place of ground truth. On KITTI, PLOT outperforms OVM3D-Det~\cite{huangtraining} by an average of $+19$ AP, and also surpasses weakly-supervised methods that rely on auxiliary sensors such as stereo~\cite{weakmono3d} and LiDAR~\cite{peng2022weakm3d}. PLOT also achieves results comparable to fully supervised baselines and open-set detectors. Compared to the previous state-of-the-art method~\cite{skvrna2025monosowa}, which relies on IMU sensors and simplified shape assumptions, PLOT achieves superior AP@0.3 (up to $+9.68$) and, as illustrated in Fig.~\ref{fig:kitti}, yields more accurate 3D boxes by leveraging trajectory-guided shape fusion and global object association. On KITTI-360, PLOT also improves over recent pseudo-labeling methods~\cite{liu2024vsrd, skvrna2025monosowa}, achieving the best AP@0.3 in both Easy and Hard regimes. 

\subsubsection{KITTI (Pedestrian).} 
Beyond the commonly evaluated `Car' category, we further assess performance on the more challenging `Pedestrian' class in the KITTI validation split (Tab.~\ref{tab:kitti_ped}). Pedestrians introduce additional challenges due to their small size, frequent occlusions, and non-rigid motion, which often violate shape assumptions adopted by existing methods~\cite{huangtraining, skvrna2025monosowa}. Despite these difficulties—and although pedestrian evaluation is less commonly reported in prior pseudo-labeling works—PLOT achieves strong performance and surpasses existing approaches. This result indicates that trajectory-guided shape fusion remains effective even when rigid shape assumptions break down, enabling robust localization and orientation estimation for small and dynamic objects.


\subsubsection{Waymo-Open (Vehicle).}  
\begin{figure*}[t]
\centering
\begin{minipage}[bt]{0.59\linewidth}
    \centering
    \begin{tabular}{>{\centering\arraybackslash}p{0.42\linewidth}
                >{\centering\arraybackslash}p{0.4\linewidth}}
    \scriptsize MonoSOWA~\cite{skvrna2025monosowa} & \scriptsize \textbf{PLOT (Ours)} \\
\end{tabular}
    \includegraphics[width=\linewidth]{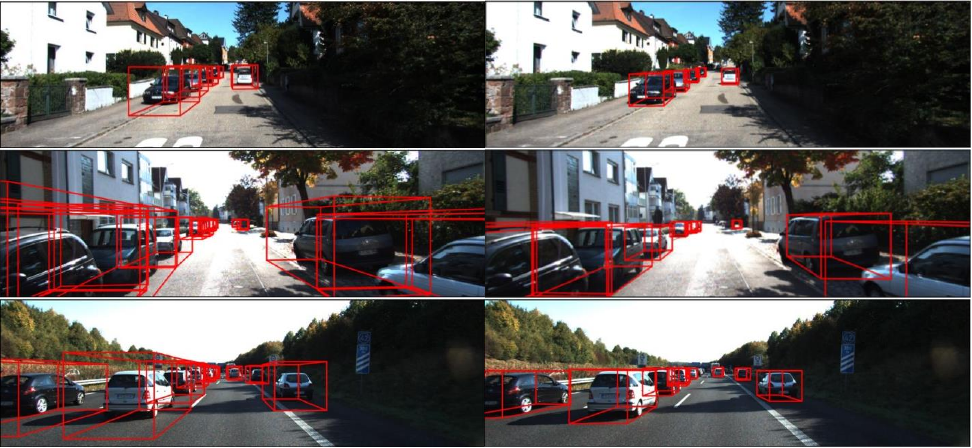}
    \caption{Qualitative comparisons on KITTI. Only the \fcolorbox{red}{white}{predicted} boxes are displayed for clarity.}
    \label{fig:kitti}
\end{minipage}\hfill
\begin{minipage}[bt]{0.39\linewidth}
    \centering
    \begin{tabular}{>{\centering\arraybackslash}p{0.4\linewidth}
                >{\centering\arraybackslash}p{0.4\linewidth}}
    \scriptsize 3D-MOOD~\cite{yang20253d} & \scriptsize \textbf{PLOT (Ours)} \\
\end{tabular}
    \includegraphics[width=\linewidth]{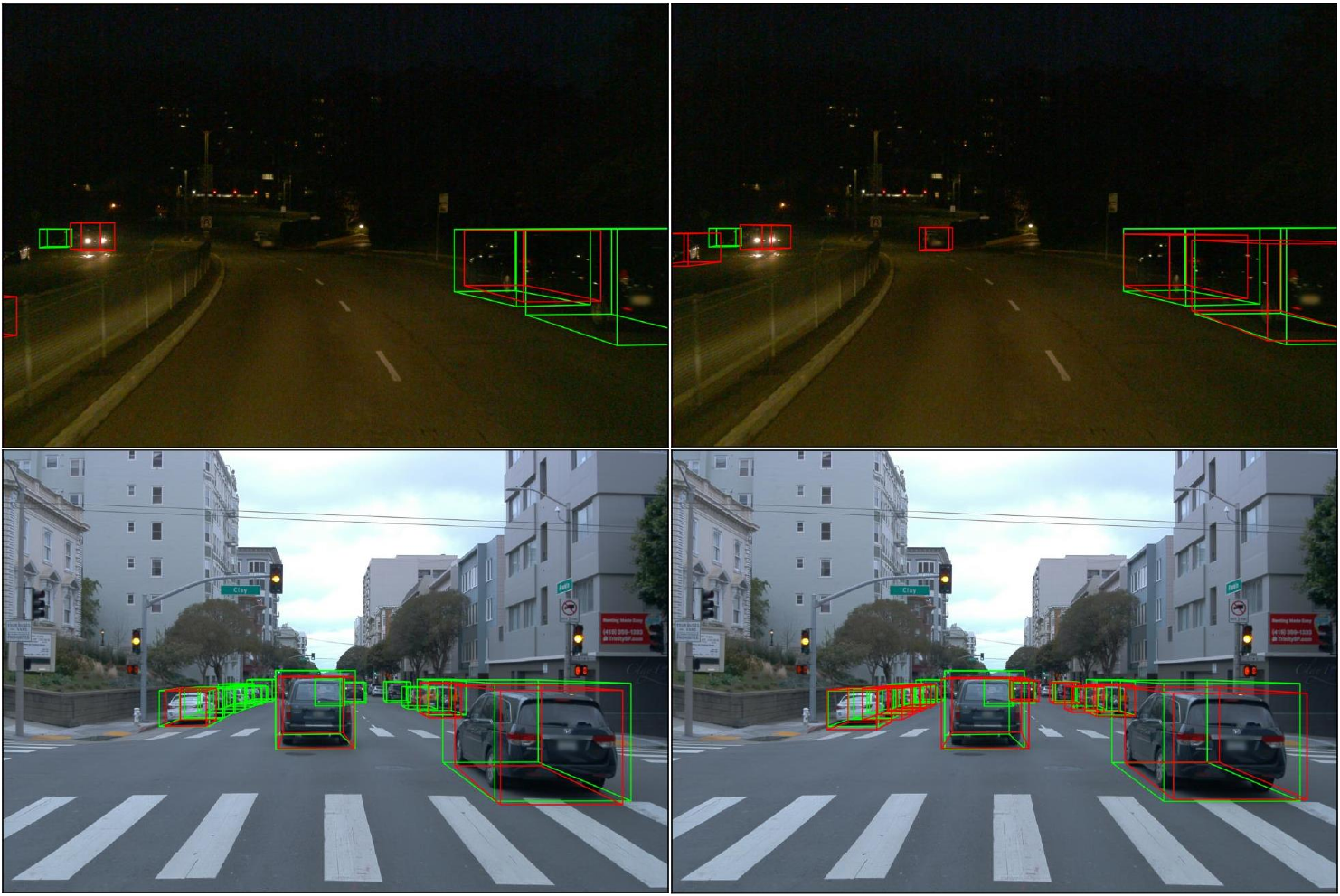}
    \caption{Qualitative comparison on Waymo-Open.}
    \label{fig:waymo}
\end{minipage}
\end{figure*}

We report results on the Waymo-Open dataset in Tab.~\ref{tab:waymo}. Unlike KITTI benchmarks, Waymo features long, diverse sequences under adverse conditions such as night and rain, making it a strong proxy for real-world complexity. PLOT achieves the best performance in distant regions ($>$30m), surpassing fully supervised baselines, and shows superior AP$_\text{BEV}$ across nearly all ranges. As shown in Fig.~\ref{fig:waymo}, PLOT remains effective under challenging conditions such as night scenes, even recovering valid instances missing from the ground truth ({\color{red}red}: predictions, {\color{green}green}: ground truth), underscoring the scalability and reliability of our framework in challenging environments.

\begin{figure*}[t]
    \centering
    \captionsetup{font={small}, skip=8pt}
    \begin{tabular}{>{\centering\arraybackslash}p{0.3\textwidth}>{\centering\arraybackslash}p{0.32\textwidth}>{\centering\arraybackslash}p{0.3\textwidth}}
         \scriptsize OVM3D-Det~\cite{huangtraining} & \scriptsize 3D-MOOD~\cite{yang20253d} & \scriptsize \textbf{PLOT~(Ours})\\
    \end{tabular}
    \includegraphics[width=1.\textwidth]{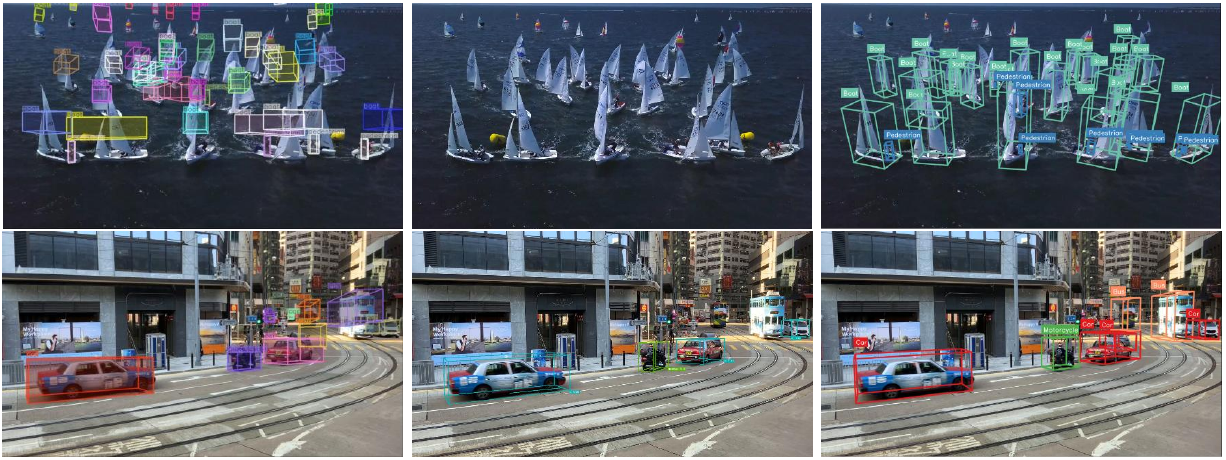}
    \caption{Qualitative comparisons in the wild.}
    \label{fig:inthewild}
\end{figure*}

\subsection{Experiments on In-the-wild Videos}
\label{subsec:wild}
To demonstrate the open-set capability of our method, we present qualitative comparisons on in-the-wild videos using identical text prompts (e.g., ``boat”) as inputs for PLOT, a state-of-the-art open-set method, 3D-MOOD~\cite{yang20253d}, and a single-image pseudo-labeling method OVM3D-Det~\cite{huangtraining}; MonoSOWA~\cite{skvrna2025monosowa}, which depends on auxiliary sensors and object shape priors, is not applicable here. As shown in Fig.~\ref{fig:cubercnn_results} and Fig.~\ref{fig:inthewild}, PLOT generalizes well under diverse and challenging conditions~(e.g., variations in scene layouts, camera setups) while maintaining accurate object attribute estimates. 
In contrast, 3D-MOOD does not detect any objects in the `boat' scene~(top) in Fig.~\ref{fig:inthewild}, and OVM3D-Det often produces noisy estimates with most falls back to priors of LLM-driven fixed objects, due to single image observations.
Such in-the-wild evaluations are particularly important, as they go beyond structured driving benchmarks and validate the scalability of our method to unconstrained real-world settings. Additional qualitative results along with video demonstrations are provided in the supplementary material.

\begin{table*}[t] \centering
\begin{minipage}[t]{0.50\linewidth}
    \caption{Ablation on trajectory-guided shape fusion and number of adj. frames.}
    \label{tab:plc_ablation}
    \resizebox{\textwidth}{!}{
    \large
    {\input{tab/tab4_v2}}
    }
\end{minipage}\hfill
    \begin{minipage}[t]{0.48\linewidth}
    \caption{Ablation on global object memory and PLOT as standalone detector.}
    \label{tab:goam_standalone}
    \resizebox{\textwidth}{!}{
    \large
    {\input{tab/tab5_gom_standalone}}}
    \begin{minipage}[t]{0.5\linewidth}
        
    \end{minipage}
\end{minipage}
\end{table*}

\subsection{Ablation Studies}

\begin{figure*}[t]
    \centering
    \includegraphics[width=\textwidth]{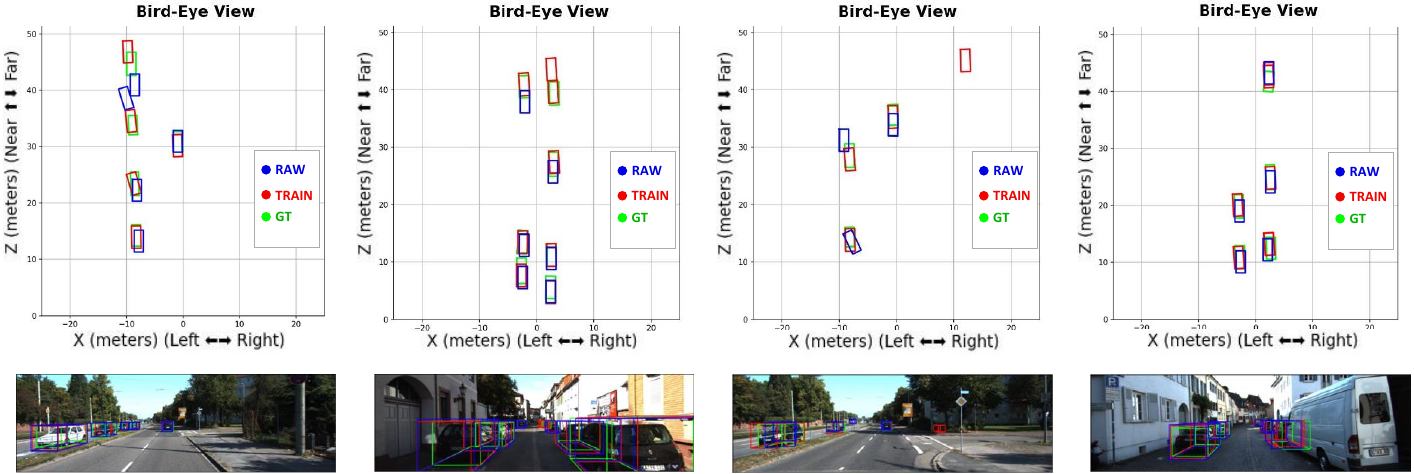}
    \caption{BEV comparison between the detection results of raw pseudo-labels and the detector trained with it, on KITTI~\cite{kitti}.}
    \label{fig:raw_trained_qual}
\end{figure*}

\begin{table*}[t]
    \centering
    \caption{Quantitative analysis of the pseudo-label accuracy for the `Car' class on the KITTI~\cite{kitti} validation set.}
    \label{tab:labelacc}
    \resizebox{0.9\linewidth}{!}
    {\input{tab/tab_labelacc}}
\end{table*}

\subsubsection{Efficacy of Trajectory-guided Shape Fusion.} 
Tab.~\ref{tab:plc_ablation} compares trajectory-guided shape fusion with varying frame counts, reporting detection performance and labeling time per frame on KITTI. Performance improves with more frames, achieving up to $1.7\times$ higher average AP compared to single-frame labels. 
The rightmost column of the table shows the computation time of the whole pipeline, which grows proportionally with the number of objects, yet remains practical~(running at $2.83$s per frame on a single CPU with a RTX 3090 GPU) with 20 objects per scene, while the KITTI dataset on average contains about 10–12 objects per frame. 
Considering both accuracy and computation time, we adopt a 20-frame window for the experiments.

\subsubsection{PLOT as Standalone 3D Detector.} 
In the top 2 rows of Tab.~\ref{tab:goam_standalone}, we examine the feasibility of using PLOT as a standalone 3D detector without training an additional M3OD model, MonoDETR~\cite{zhang2023monodetr} on KITTI. Remarkably, the raw pseudo-labels already surpass the previous pseudo-labeling method~(OVM3D-Det~\cite{huangtraining}) and the weakly-supervised method~(MonoGRNet~\cite{qin2021monogrnet})~(see Tab.~\ref{tab:kitti}), suggesting the utility of PLOT in scenarios where training is infeasible, such as in-the-wild videos or cases with only a single sequence available. However, the integration of uncertainty-aware depth estimation and 2D–3D geometric consistency in MonoDETR mitigates depth errors and label inconsistencies, yielding more stable and accurate 3D detections. 
Fig.~\ref{fig:raw_trained_qual} provides qualitative evidence supporting this observation by comparing ground truth, raw pseudo-labels, and the predictions obtained after training MonoDETR on the generated labels in BEV space. While the raw pseudo-labels already exhibit reasonable localization quality, depth noise occasionally shifts object centers (noticeable in far objects); training reduces these deviations through regularization, resulting in more consistent geometric alignment.

\subsubsection{Necessity of Global Object Memory.} 
We evaluated the impact of global object memory on KITTI-360~(K360) in the bottom 2   rows of Tab.~\ref{tab:goam_standalone}, as KITTI’s 3D detection benchmark (single-frame) is not suitable for assessing GOM, whereas KITTI-360 provides continuous sequences. The capability of GOM to reduce 2D detector-induced noise and maintain temporal consistency results in more stable and accurate pseudo-labels. Such video-consistent label generation is directly relevant for practical annotation pipelines and downstream 3D perception tasks. 
In the supplementary material, we also provide video-consistent label results on the Waymo dataset as well as the breakdown of temporal consistent 3D labels on in-the-wild scene in the supplementary material.

\subsubsection{Analysis of Pseudo-Labels Accuracy.} 
Tab.~\ref{tab:labelacc} compares raw pseudo-labels from PLOT and OVM3D-Det~\cite{huangtraining} using nuScenes~\cite{nuscenes} metrics without any training. 
PLOT yields a higher detection rate due to its label improvement scheme, which minimizes missing and duplicate detections.
The AOE (Average Orientation Error) difference shows that PLOT improves over OVM3D-Det (which also uses PCA) on both all and static objects only, indicating that our pseudo-LiDAR completion provides more reliable geometry, even when only limited visible surfaces are available, as also qualitatively shown for parked cars in Fig.~A6 of the supplementary material.
The improved geometry also benefits the object size estimation, evidenced by a significant drop in ASE (Average Scale Error).

%% file: tab/tab1_kitti.tex
\begin{tabular}{c|l|c|c|*{3}{c}|*{3}{c}|*{3}{c}|*{3}{c}}
\toprule
& \multirow{2}{*}{Method} & \multirow{2}{*}{Labels} & \multirow{2}{*}{Extra} & \multicolumn{3}{c|}{ $\mathrm{AP_{3D}@IoU=0.3}$$\uparrow$} & \multicolumn{3}{c|}{$\mathrm{AP_{BEV}@IoU=0.3}$$\uparrow$}  & \multicolumn{3}{c|}{ $\mathrm{AP_{3D}@IoU=0.5}$$\uparrow$} & \multicolumn{3}{c}{$\mathrm{AP_{BEV}@IoU=0.5}$$\uparrow$} \\ 
& & & & Easy & Mod. & Hard & Easy & Mod. & Hard & Easy & Mod. & Hard & Easy & Mod. & Hard  \\
\midrule
\multirow{2}{*}{\rotatebox{90}{Sup.}} & \fade{DEVIANT~\cite{kumar2022deviant}} & \fade{GT-3D} & \fade{\xmark} & \fade{-} & \fade{-} & \fade{-} & \fade{-} & \fade{-} & \fade{-} & \fade{61.00} & \fade{46.00} & \fade{40.18} & \fade{65.28} & \fade{49.63} & \fade{43.50} \\
& \fade{MonoDETR~\cite{zhang2023monodetr}} & \fade{GT-3D} & \fade{\xmark} & \fade{79.72} & \fade{65.87} & \fade{58.83} & \fade{80.30} & \fade{67.16} & \fade{59.54} & \fade{68.05} & \fade{48.42} & \fade{43.48} & \fade{72.34} & \fade{51.97} & \fade{46.94} \\
\midrule
\multirow{2}{*}{\rotatebox{90}{Opn.}} & \fade{OVMono3D~\cite{yao2024open}} & \fade{GT-3D} & \fade{\xmark} & \fade{74.01} & \fade{51.25} & \fade{42.40} & \fade{76.93} & \fade{53.85} & \fade{44.89} & \fade{51.23} & \fade{33.09} & \fade{27.24} & \fade{55.56} & \fade{36.47} & \fade{30.25} \\
 & \fade{3D-MOOD~\cite{yang20253d}} & \fade{GT-3D} & \fade{Depth} & \fade{81.97} & \fade{64.16} & \fade{54.36} & \fade{83.04} & \fade{66.69} & \fade{56.79} & \fade{60.74} & \fade{43.81} & \fade{36.95} & \fade{64.93} & \fade{47.22} & \fade{40.00} \\
\midrule
\midrule
\multirow{4}{*}{\rotatebox{90}{Weak}} & MonoGRNet~\cite{qin2021monogrnet} & GT-2D & Attr-Net & 56.16 & 42.61 & 35.36 & 58.61 & 48.75 & 41.49 & 25.66 & 21.57 & 17.40 & 32.23 & 26.88 & 22.47 \\
& WeakM3D~\cite{peng2022weakm3d} & GT-2D & LiDAR & \U{78.44} & \U{56.42} & 45.81 & \U{81.17} & \U{59.87} & \U{48.98} & 50.16 & 29.94 & 23.11 & 58.20 & 38.02 & 30.17 \\
& WeakMono3D~\cite{weakmono3d} &  GT-2D & Stereo & - & - & - & - & - & - & 49.37 & \U{39.01} & \U{36.34} & 54.32 & 42.83 & \U{40.07} \\
& SKD-WM3D~\cite{jiang2024weaklysingle} & GT-2D & Depth${\dagger}$ & - & - & - & - & - & - &  50.21 & \B 41.57 & \B 36.92 & 55.47 & \B 44.35 & \B 41.86 \\
\midrule
\multirow{3}{*}{\rotatebox{90}{Pseudo}} & OVM3D-Det~\cite{huangtraining} & GSAM & GPT-4 & 44.48 & 33.29 & 26.69 & 47.42 & 35.96 & 27.99 & 25.63 & 18.85 & 15.67 & 34.52 & 24.46 & 20.40 \\
& MonoSOWA~\cite{skvrna2025monosowa} & MViT2 & VPS & 72.70 & 56.30 & \U{47.70} & 73.38 & 57.23 & 48.59 & \U{51.55} & 37.09 & 33.15 & \U{59.76} & \U{44.08} & 36.99 \\
& \textbf{PLOT (Ours)} & GSAM & Video & \B 80.48 & \B 60.83 & \B 51.49 & \B 83.06 & \B 63.59 & \B 54.15 & \B 52.02 & 36.78 & 30.40 & \B 60.25 & 42.80 & 35.77 \\
\bottomrule
\end{tabular}

%% file: tab/tab2_kitti360.tex
\begin{tabular}{c|l|c|c|*{2}{c}|*{2}{c}|*{2}{c}|*{2}{c}}
\toprule
& \multirow{2}{*}{Method} & \multirow{2}{*}{Labels} & \multirow{2}{*}{Extra} & \multicolumn{2}{c|}{ $\mathrm{AP_{3D}@0.3}$$\uparrow$} & \multicolumn{2}{c|}{$\mathrm{AP_{BEV}@0.3}$$\uparrow$}  & \multicolumn{2}{c|}{ $\mathrm{AP_{3D}@0.5}$$\uparrow$} & \multicolumn{2}{c}{$\mathrm{AP_{BEV}@0.5}$$\uparrow$} \\ 
 & & & & Easy & Hard & Easy & Hard & Easy & Hard & Easy & Hard  \\ 
\midrule
\multirow{1}{*}{Sup.} & \fade{MonoDETR~\cite{zhang2023monodetr}} & \fade{GT-3D} & \fade{\xmark} & \fade{64.32} & \fade{57.83} & \fade{66.42} & \fade{60.31} & \fade{47.69} & \fade{38.76} & \fade{52.76} & \fade{43.56} \\
\midrule
\multirow{2}{*}{Open.} & OVMono3D~\cite{yao2024open} & Zero-shot & \xmark & 41.82 & 31.94 & 49.45 & 39.08 & 4.49 & 3.39 & 29.18 & 22.07 \\
 & 3D-MOOD~\cite{yang20253d} & Zero-shot & \xmark & 21.15 & 13.12 & 32.68 & 21.41 & 0.06 & 0.04 & 7.62 & 4.59 \\
\midrule
\multirow{2}{*}{Weak.} & WeakM3D~\cite{peng2022weakm3d} & GT-2D & LiDAR & 21.25 & 15.34 & 29.89 & 24.01 & 2.96 & 2.01 & 8.10 & 2.96 \\
& Autolabels~\cite{zakharov2020autolabeling} & GT-2D & LiDAR, Shape & 12.92 & 9.94 & 48.16 & 37.34 & 4.69 & 2.79 & 20.18 & 14.33 \\
\midrule
\multirow{3}{*}{Pseudo.} & VSRD~\cite{liu2024vsrd} & GT-2D & Pose & \U{50.86} & 43.45 & \U{58.40} & \U{50.61} & 21.77 & 16.46 & 29.07 & 22.83 \\
& MonoSOWA~\cite{skvrna2025monosowa} & MViT2 & VPS & 42.72 & \U{46.59} & 50.84 & 49.22 & \B 29.98 & \B 27.56 & \B 38.41 & \B 35.26 \\
& \textbf{PLOT (Ours)} & GSAM & Video & \B 54.78 & \B 48.75 & \B 60.75 & \B 54.58 & \U{27.34} & \U{21.98} & \U{36.88} & \U{30.45} \\
\bottomrule
\end{tabular}

%% file: tab/atab1_kitti_ped.tex
\begin{tabular}{c|l|c|c|*{3}{c}|*{3}{c}|*{3}{c}|*{3}{c}}
\toprule
& \multirow{2}{*}{Method} & \multirow{2}{*}{Labels} & \multirow{2}{*}{Extra} & \multicolumn{3}{c|}{ $\mathrm{AP_{3D}@IoU=0.3}$$\uparrow$} & \multicolumn{3}{c|}{$\mathrm{AP_{BEV}@IoU=0.3}$$\uparrow$}  & \multicolumn{3}{c|}{ $\mathrm{AP_{3D}@IoU=0.5}$$\uparrow$} & \multicolumn{3}{c}{$\mathrm{AP_{BEV}@IoU=0.5}$$\uparrow$} \\ 
& & & & Easy & Mod. & Hard & Easy & Mod. & Hard & Easy & Mod. & Hard & Easy & Mod. & Hard  \\
\midrule
\multirow{4}{*}{\rotatebox{90}{Sup.}} & \fade{MonoDIS~\cite{simonelli2019disentangling}} & \fade{GT-3D} & \fade{\xmark} & \fade{-} & \fade{-} & \fade{-} & \fade{-} & \fade{-} & \fade{-} & \fade{3.20} & \fade{2.28} & \fade{1.71} & \fade{4.04} & \fade{3.19} & \fade{2.45} \\
& \fade{MonoXiver~\cite{liu2023denoising}} & \fade{GT-3D} & \fade{\xmark} & \fade{-} & \fade{-} & \fade{-} & \fade{-} & \fade{-} & \fade{-} & \fade{7.95} & \fade{5.49} & \fade{4.62} & \fade{-} & \fade{-} & \fade{-} \\
& \fade{GUP-Net~\cite{gupnet}} & \fade{GT-3D} & \fade{\xmark} & \fade{-} & \fade{-} & \fade{-} & \fade{-} & \fade{-} & \fade{-} & \fade{9.37} & \fade{6.84} & \fade{5.73} & \fade{-} & \fade{-} & \fade{-} \\
& \fade{DEVIANT~\cite{kumar2022deviant}} & \fade{GT-3D} & \fade{\xmark} & \fade{-} & \fade{-} & \fade{-} & \fade{-} & \fade{-} & \fade{-} & \fade{9.85} & \fade{7.18} & \fade{5.42} & \fade{-} & \fade{-} & \fade{-} \\
\midrule
\multirow{1}{*}{\rotatebox{90}{W}} & WeakM3D~\cite{peng2022weakm3d} & GT-2D & LiDAR & - & - & - & 3.79 & 3.21 & 3.12 & - & - & - & - & - & - \\
\midrule
\multirow{2}{*}{\rotatebox{90}{Psd.}} & OVM3D-Det~\cite{huangtraining} & GSAM & GPT-4 & 10.63 & 8.96 & 7.32 & 11.25 & 9.36 & 7.85 & 8.19 & 6.88 & 5.56 & 9.11 & 7.71 & 6.27 \\
& \textbf{PLOT (Ours)} & GSAM & Video & \B 16.64 & \B 14.39 & \B 12.27 & \B 17.66 & \B 15.01 & \B 12.76 & \B 13.97 & \B 11.71 & \B 9.84 & \B 14.52 & \B 12.48 & \B 10.59 \\
\bottomrule
\end{tabular}

%% file: tab/tab3_waymo.tex
\begin{tabular}{c|l|c|c|*{4}{c}|*{4}{c}}
\toprule
& \multirow{2}{*}{Method} & \multirow{2}{*}{Labels} & \multirow{2}{*}{Extra} & \multicolumn{4}{c|}{ $\mathrm{AP_{3D}@0.5}$$\uparrow$} & \multicolumn{4}{c}{$\mathrm{AP_{BEV}@0.5}$$\uparrow$} \\ 
 & & & & All & 0-30m & 30-50m & 50-$\infty$m & All & 0-30m & 30-50m & 50-$\infty$m  \\ 
\midrule
\multirow{3}{*}{Sup.} & \fade{DEVIANT~\cite{kumar2022deviant}} & \fade{GT-3D} & \fade{\xmark} & \fade{10.29} & \fade{26.75} & \fade{4.95} & \fade{0.16} & \fade{-} & \fade{-} & \fade{-} & \fade{-} \\
& \fade{CaDDN~\cite{reading2021categorical}} & \fade{GT-3D} & \fade{LiDAR} & \fade{16.51} & \fade{44.87} & \fade{8.99} & \fade{0.58} & \fade{-} & \fade{-} & \fade{-} & \fade{-} \\
& \fade{MonoDETR~\cite{zhang2023monodetr}} & \fade{GT-3D} & \fade{\xmark} & \fade{21.41} & \fade{37.89} & \fade{18.61} & \fade{3.69} & \fade{23.63} & \fade{39.10} & \fade{20.95} & \fade{4.70} \\
\midrule
\multirow{2}{*}{Open.} & OVMono3D~\cite{yao2024open} & Zero-shot & \xmark & 5.46 & 14.68 & 5.61 & 0.65 & 6.11 & 15.24 & 6.54 & 0.99 \\
& 3D-MOOD~\cite{yang20253d} & Zero-shot & \xmark & 1.13 & 2.44 & 1.68 & 0.05 & 4.60 & 9.80 & 6.71 & 0.30 \\
\midrule
\multirow{1}{*}{Weak.} & WeakM3D~\cite{peng2022weakm3d} & GT-2D & LiDAR & 4.50 & 12.16 & 3.67 & 0.40 & 8.18 & 20.31 & 7.50 & 0.99 \\
\midrule
\multirow{2}{*}{Pseudo.} & MonoSOWA~\cite{skvrna2025monosowa} & MViT2 & VPS & \U{13.46} & \U{24.65} & \U{5.87} & \U{4.55} & \U{18.98} & \U{33.51} & \U{12.02} & \U{4.55} \\
& \textbf{PLOT (Ours)} & GSAM & Video & \B 14.04 & \B 26.97 & \B 17.71 & \B 4.87 & \B 23.32 & \B 41.74 & \B 31.39 & \B 8.22 \\
\bottomrule
\end{tabular}

%% file: tab/tab4_v2.tex
\begin{tabular}{l|*{3}{c}|*{3}{c}|c}
\toprule
\multirow{2}{*}{\#Fr.} & \multicolumn{3}{c|}{ $\mathrm{AP_{3D}@IoU=0.3}$$\uparrow$} & \multicolumn{3}{c|}{$\mathrm{AP_{BEV}@IoU=0.3}$$\uparrow$} & Time \\ 
 & Easy & Mod. & Hard & Easy & Mod. & Hard & ($10/20$ obj.) \\ 
\midrule
1 & 51.75 & 37.48 & 30.63 & 57.41 & 41.96 & 34.00 & \textbf{0.46} / \textbf{0.47} \\
+2 & 62.41 & 48.06 & 40.31 & 67.57 & 51.39 & 43.35 & 0.66 / 0.70 \\
+8 & 77.23 & 56.16 & 48.43 & 80.27 & 59.13 & 51.29 & 1.56 / 1.73 \\
+20 & \B 80.48 & \B 60.83 & \B 51.49 & \B 83.06 & \B 63.59 & \B 54.15 & 2.45 / 2.83 \\
\bottomrule
\end{tabular}

%% file: tab/tab5_gom_standalone.tex
\begin{tabular}{c|c|c|*{3}{c}|*{3}{c}}
\toprule
& \multirow{2}{*}{Train} & \multirow{2}{*}{GOM} & \multicolumn{3}{c|}{ $\mathrm{AP_{3D}@IoU=0.3}$$\uparrow$} & \multicolumn{3}{c}{$\mathrm{AP_{BEV}@IoU=0.3}$$\uparrow$}  \\ 
 & & & Easy & Mod. & Hard & Easy & Mod. & Hard \\ 
\midrule
\multirow{2}{*}{\rotatebox{90}{KIT.}} & \xmark & \checkmark & 59.82 & 51.00 & 45.40 & 62.41 & 54.18 & 48.44 \\
& \checkmark & \checkmark & \B 80.48 & \B 60.83 & \B 51.49 & \B 83.06 & \B 63.59 & \B 54.15 \\
\midrule
\midrule
\multirow{2}{*}{\rotatebox{90}{K360}} &\checkmark & \xmark & 48.14 & - & 40.28 & 54.86 & - & 46.61 \\
& \checkmark & \checkmark & \B 54.78 & - & \B 48.75 & \B 60.75 & - & \B 54.58 \\
\bottomrule
\end{tabular}

%% file: tab/tab_labelacc.tex
\begin{tabular}{c|c|*{3}{c}|*{3}{c}|*{3}{c}}
\toprule
 \multirow{2}{*}{Method} & Object & \multicolumn{3}{c|}{$\mathrm{AP_{2D}}$$\uparrow$ (\%)} & \multicolumn{3}{c|}{AOE$\downarrow$ (rad)} & \multicolumn{3}{c}{ASE$\downarrow$ (1-IoU)} \\ 
  & Type & Near & Mid. & Far & Near & Mid. & Far & Near & Mid. & Far \\ 
\midrule
OVM3D-Det~\cite{huangtraining} & static & 64.43 & 49.87 & 40.92 & 1.245 & 1.200 & 1.380 & 0.338 & 0.312 & 0.412 \\
\textbf{PLOT (Ours)} & static & \textbf{80.40} & \textbf{73.03} & \textbf{64.18} & \textbf{0.716} & \textbf{0.695} & \textbf{0.794} & \textbf{0.149} & \textbf{0.140 }& \textbf{0.249} \\  
\midrule
OVM3D-Det~\cite{huangtraining} & All & 65.57 & 59.68 & 48.27 & 1.360 & 1.127 & 1.413 & 0.329 & 0.297 & 0.318 \\ 
\textbf{PLOT (Ours)} & All & \textbf{83.98} & \textbf{79.40} & \textbf{71.53} & \textbf{0.536} &\textbf{ 0.621} & \textbf{0.712} & \textbf{0.136} & \textbf{0.131} & \textbf{0.168 }\\
\bottomrule
\end{tabular}

%% file: sec/5_conclusion.tex
\section{Conclusion}
In this paper, we introduced PLOT, a framework for generating reliable 3D annotations from monocular videos through tracking-driven object association and label refinement. Beyond monocular 3D detection, this video-based formulation holds potential for broader 3D tasks that suffer from missing camera information or incomplete shapes, such as CAD model retrieval and object-level reconstruction. We hope that this paradigm provides a scalable alternative to data-intensive monocular 3D detection pipelines and opens new directions for video-based 3D understanding without explicit supervision.


\subsubsection{Limitations.}
Like most open-set 3D perception frameworks, PLOT relies on pre-trained depth estimators. While this dependency is not unique to our approach, depth errors under noise or at long ranges remain a key bottleneck. In practice, however, training downstream M3OD models on the generated pseudo-labels helps reduce these errors by regularizing noisy depth estimates. In addition, orientation for static objects remains inherently ambiguous under partial-view observations. While we adopt PCA-based estimation, the trajectory-guided shape completion in PLOT provides more complete object geometry over time, making such estimators substantially more robust in practice. Further analysis and additional failure cases are provided in the supplementary materials.

%% file: sec/6_appendix.tex
In this supplementary material, we first describe the details of the method required to reproduce our approach in Sec. \ref{appendix:method}, followed by the experimental details needed for comparisons with baselines in Sec. \ref{appendix:setup}. 
We then provide additional experiments in Sec. \ref{appendix:exp}, and present a dedicated analysis of failure cases in Sec. \ref{appendix:failure}, which summarizes the main limitations and discusses remaining challenges.
Finally, we show qualitative results that highlight aspects of our approach not fully reflected by quantitative metrics in Sec. \ref{appendix:qual}.

In addition to this text supplement, our supplementary material includes two additional components: (i) partial source code for reproducibility, and (ii) video-consistent demos. 
The demos include several in-the-wild videos, which illustrate the frame-to-frame consistency of pseudo-labeling without training, and video demonstrations on some of the Waymo~\cite{waymo} scenes, which visualize predictions from a model trained with our pseudo-labels. 

\section{Method Details}
\label{appendix:method}
\subsection{Usage of Foundation Models} 
Following recent literature on generalizable perception pipelines~\cite{huangtraining, som2024}, we leverage several foundation models for pseudo-label generation: Grounded-SAM~\cite{ren2024grounded} for 2D detection and segmentation, AllTracker~\cite{harley2025alltracker} for dense point tracking, and UniDepth~\cite{unidepth} or MoGe2~\cite{wang2025moge} for metric depth estimation. 
For 2D label generation with Grounded-SAM~\cite{ren2024grounded}, we use the prompts with the categories: `Car'/`Vehicle' and `Pedestrian' for driving benchmarks and adapt relevant class labels for in-the-wild videos, with a box threshold of $0.3$ and a text threshold of $0.25$. To ensure fairness, we adopt the same text prompts as used in other prompt-based baselines~\cite{yao2024open, yang20253d, huangtraining}. 
For depth estimation, we use ground-truth camera intrinsics on driving benchmarks and estimated intrinsics from each depth model for in-the-wild videos. We find UniDepth~\cite{unidepth} to be highly accurate in driving scenarios, and therefore adopt it for experiments on driving benchmarks, while MoGe2~\cite{wang2025moge} is used for in-the-wild videos. 
Our choice of 2D detector and depth estimator follows OVM3D-Det~\cite{huangtraining}, where as MonoSOWA~\cite{skvrna2025monosowa} uses COCO-trained MViT-v2~\cite{mvit2} detector and Metric3Dv2~\cite{metric3dv2} depth estimator.

\subsection{Hyperparameters for Pseudo-labeling} 
Pseudo-label generation uses parameters chosen based on typical object persistence and motion patterns observed in general video sequences. Pseudo-labels are generated on 20-frame tracking windows (see Tab.~5, main paper), where dense tracks are computed jointly across frames; longer sequences in KITTI-360~\cite{kitti360} and Waymo~\cite{waymo} are split into non-overlapping windows of this length. Object visibility for camera-motion estimation (Sec.~3.2; Eq.~(4)) is computed from point-track visibility with a threshold of $0.6$. 
Objects unmatched for $4$ consecutive frames are considered to have left the scene, and tracklets shorter than $5$ frames are discarded. The IoU threshold between a tracklet and its corresponding detection in Hungarian matching~\cite{kuhn1955hungarian} (Eq.~(2), Sec.~3.1) is set to $0.4$. During camera motion estimation, points with depth greater than $50$ m are filtered following common practice in 3D perception to suppress noisy estimates, while object motion estimation uses all available points. An object is considered moving if its displacement $|| \hat{\mathbf{p}}_i^s - \hat{\mathbf{p}}_i^t ||$ exceeds $2$ m. 
In scenarios with substantially different temporal sampling rates, temporal thresholds may require adjustment.

\subsection{MonoDETR Training} 
Using the pseudo-labels generated on the training split, we train MonoDETR~\cite{zhang2023monodetr}, a monocular 3D detector, in a fully supervised manner. Its performance is then evaluated on the corresponding validation or test split, depending on the dataset. All training hyperparameters follow the official setting without modification.


\section{Experimental Details}
\label{appendix:setup}

\subsection{Hardware Specification} 
Pseudo-label generation and the run-time analysis reported in Tab.~5 of the main paper are conducted on a PC equipped with an Intel Core i9-14900KF CPU and a single NVIDIA RTX-3090 GPU. The training of MonoDETR~\cite{zhang2023monodetr} is performed separately on a PC with a single NVIDIA A100 GPU.

\subsection{Benchmark Datasets}
We evaluate PLOT on three monocular 3D object detection~(M3OD) benchmarks with video sequences: KITTI~\cite{kitti}, KITTI-360~\cite{kitti360}, and Waymo-Open~\cite{waymo}. For KITTI, we adopt the official training and validation split of the 3D object detection benchmark, consisting of 3,712 training images and 3,769 validation images. Since the KITTI M3OD benchmark is defined for single images, we leverage adjacent frames from the raw video sequences for each labeled image to enable temporal reasoning, simliar to the previous work~\cite{skvrna2025monosowa}. For KITTI-360, we follow the protocol in VSRD~\cite{liu2024vsrd}, which uses 6 training sequences with 44,178 frames and 1 test sequence with 2,459 frames. 
For the Waymo-Open dataset, we follow the official split, consisting of 798 training sequences (158,080 frames) and 202 validation sequences (39,988 frames). 

\subsection{In-the-wild Videos}
Since all public M3OD benchmarks are limited to driving scenes, we further evaluate generalization by applying PLOT to diverse monocular videos outside this domain. We specifically select datasets that capture scenes rarely covered by standard 3D detection benchmarks, including surveillance footage (MOT17~\cite{MOT16}, DIVOTrack~\cite{hao2024divotrack}), handheld recordings (MOSE~\cite{ding2023mose}), and crowd-sourced videos (GMOT-40~\cite{bai2021gmot}, Pexels~\cite{pexels2025}). Unlike structured driving environments with forward-facing cameras and narrow fields of view, these in-the-wild settings introduce unconstrained camera motions, varied scene structures, and diverse object categories. Such characteristics make them particularly valuable for assessing robustness under open-set evaluation, where models must handle novel environments without domain-specific assumptions. To illustrate this, we compare results on representative scenarios, including a CCTV-like view (Fig. 2, main paper) and a `boat' scene (Fig.7, main paper). In these experiments, we use raw pseudo-labels without training and compare them with prior pseudo-labeling methods\cite{huangtraining} as well as zero-shot predictions from open-set detectors\cite{yang20253d}, using the same object categories for fairness.

\subsection{Baseline Methods}
We compare PLOT with three categories of baselines: weakly-supervised M3OD models, open-set detectors, and previous pseudo-labeling approaches.
\subsubsection{Weakly-supervised M3OD models.} 
We compare against recent weakly-supervised methods~\cite{qin2021monogrnet, peng2022weakm3d, weakmono3d, jiang2024weaklysingle}, which avoid full 3D supervision but rely on additional priors, such as attribute estimation networks~\cite{qin2021monogrnet} or depth completion networks~\cite{jiang2024weaklysingle}, as well as auxiliary inputs (e.g., LiDAR, stereo, or depth). These methods also typically exploit 2D ground-truth annotations to compensate for missing 3D supervision. In contrast, PLOT generates pseudo-labels directly from monocular videos without requiring such priors or auxiliary inputs.

\subsubsection{Open-set M3OD models.}
Given the open-vocabulary nature of our approach, we also compare with open-set monocular 3D detectors~\cite{yao2024open, yang20253d}. These models are trained on large-scale cross-domain datasets with dense 3D annotations, such as Omni3D~\cite{brazil2023omni3d}, which already include subsets of driving benchmarks (e.g., KITTI~\cite{kitti} and nuScenes~\cite{nuscenes}). Although this overlap makes direct benchmark comparisons less informative, it highlights the practical gap between methods that rely on curated large-scale 3D supervision and PLOT, which learns directly from monocular video without additional annotations or domain-specific priors.

\subsubsection{Pseudo-labeling methods.}
We further compare against prior pseudo-labeling approaches~\cite{huangtraining, liu2024vsrd, skvrna2025monosowa}. These methods typically rely on additional priors, including LLM-estimated object sizes~\cite{huangtraining}, known sensor poses~\cite{liu2024vsrd, skvrna2025monosowa}, or fine-grained object templates tied to driving categories~\cite{skvrna2025monosowa}. In contrast, PLOT generates pseudo-labels directly from raw monocular videos without requiring such priors.

\subsection{Evaluation Metrics} 
For the KITTI~\cite{kitti} and KITTI-360~\cite{kitti360} datasets, we report average precision in 3D and bird's eye view ($\mathrm{AP_{3D}}/\mathrm{AP_{BEV}}$), computed at 40 recall positions and using two IoU thresholds: $0.3$ and $0.5$. The results are reported in three levels of difficulty: easy, moderate (KITTI only), and hard, based on the size of the object's bounding box. For the Waymo Open dataset, we evaluate on $\mathrm{Level\_2}$ objects and use the same metrics ($\mathrm{AP_{3D}}/\mathrm{AP_{BEV}}$) with an IoU threshold of $0.5$. The results are further broken down by distance ranges of objects: $[0, 30)$, $[30, 50)$, and $[50, \infty)$ meters.

\section{Additional Experiments} 
\label{appendix:exp}
All additional experiments are conducted on the KITTI~\cite{kitti} dataset. 

\begin{table}[t] 
    \caption{Ablation on the impact of tracker and depth noises.
    }
    \centering
    \resizebox{\linewidth}{!}
    {\input{tab/tab6_depth_tracker}}
    \label{tab:depth_noise_ablation} 
\end{table}

\subsection{Robustness to Model Noises}
\subsubsection{Tracking Noises.}
As part of the stage-wise analysis summarized in Tab.~\ref{tab:depth_noise_ablation}, we evaluate how variations in tracking accuracy influence downstream labeling quality. To this end, we replace the standard AllTracker~\cite{harley2025alltracker} with a tiny, less accurate variant and measure its effect on camera motion, fused geometry, and 3D attributes. As shown in the first row of Tab.~\ref{tab:depth_noise_ablation}, our method show negligible performance drops despite the reduced tracking accuracy, indicating that global object memory effectively compensates for fragmented trajectories and helps preserve temporal consistency. 

\subsubsection{Non-uniform Spatial and Temporal Depth Noises.}
Since PLOT relies on monocular depth, one may question its stability under depth noises, such as spatial noises and temporal scale fluctuations.
To assess this, we inject controlled perturbations into the estimated depth on a sampled Waymo-Open~\cite{waymo} scene and evaluate depth quality, camera motion, and labeling accuracy. 
As shown in Tab.~\ref{tab:depth_noise_ablation}, camera motion estimation remains stable due to reliable background correspondences.
Depth error grows mainly under range-dependent noise, while spatial noise causes only mild degradation in shape fusion and 3D attributes. 
In contrast, non-uniform temporal scale noise leads to larger drops in box AP, reflecting a depth sensitivity shared with other monocular-depth–based pipelines and indicating that advances in depth estimation will naturally improve PLOT.
The last two rows show results with Sim(3) alignment (introducing the optional scale term in Eq.4 and Eq.5 in main paper), which improves the box AP scores with stronger temporal noise.
Furthermore, as shown in the supplemental video and the frame-wise BEV visualization in Fig.~\ref{fig:app_wild_bev}, modern depth estimators exhibit minimal flickering or inter-frame scale drift in large-scale scenes.

\begin{table*}[bt] \centering
\begin{minipage}[t]{0.49\linewidth}
    \caption{Impact of camera motion on orientation estimation.}
    \vspace{-0.5em}
    \label{tab:orient_ablation}
    \resizebox{\textwidth}{!}{
    \large
    {\input{tab/atab2_orient}}
    }
\end{minipage}\hfill
    \begin{minipage}[t]{0.49\linewidth}
    \caption{Impact of MDE-estimated camera intrinsics.} 
    \vspace{-0.5em}
    \label{tab:intrinsic_ablation}
    \resizebox{\textwidth}{!}{
    \large
    {\input{tab/atab5_intrinsics}}
    }
\end{minipage}
\end{table*}

\subsection{Motion-guided Orientation Estimation} 
In Tab.~\ref{tab:orient_ablation}, we evaluate the impact of estimated camera motion (i.e., relative camera poses) in transforming local object trajectories into world-space alike trajectories. 
Without explicit motion reasoning to distinguish between static and dynamic objects, orientation estimation becomes less reliable, leading to a noticeable drop in performance ($-3.09$ AP in average). 

\subsection{Estimated Intrinsics} 
In Tab.~\ref{tab:intrinsic_ablation}, we evaluate PLOT using UniDepth\cite{unidepth}-estimated camera intrinsics in place of ground-truth values. 
Although intrinsic calibration is typically available in driving benchmarks, the results indicate that PLOT maintains performance comparable to estimated intrinsics, showing only minor degradation and demonstrating its suitability for deployment in unconstrained settings.

\subsection{Label Noise in KITTI} 
Since 3D bounding box annotations depend on the sparsity and distribution of LiDAR points, obtaining labels as precise as their 2D counterparts is inherently challenging. 
As shown in Fig. 5 in the main paper and Fig.~\ref{fig:app_kitti}, both 3D-MOOD~\cite{yang20253d} and PLOT occasionally produce detections that better align with visual evidence than the ground-truth provided by KITTI~\cite{kitti}, where some objects are missing or incompletely annotated. 
This not only supports our earlier finding that PLOT can serve as a reliable substitute for manual 3D labeling, but also highlights its potential for facilitating future dataset creation by bootstrapping new benchmarks with reduced human annotation effort.
Although KITTI remains a valuable benchmark, certain limitations in its labels can lead to an underestimation of the method's performance.

\begin{figure*}[bt]
\centering
\begin{minipage}[bt]{0.70\linewidth}
    \centering
    \includegraphics[width=\linewidth]{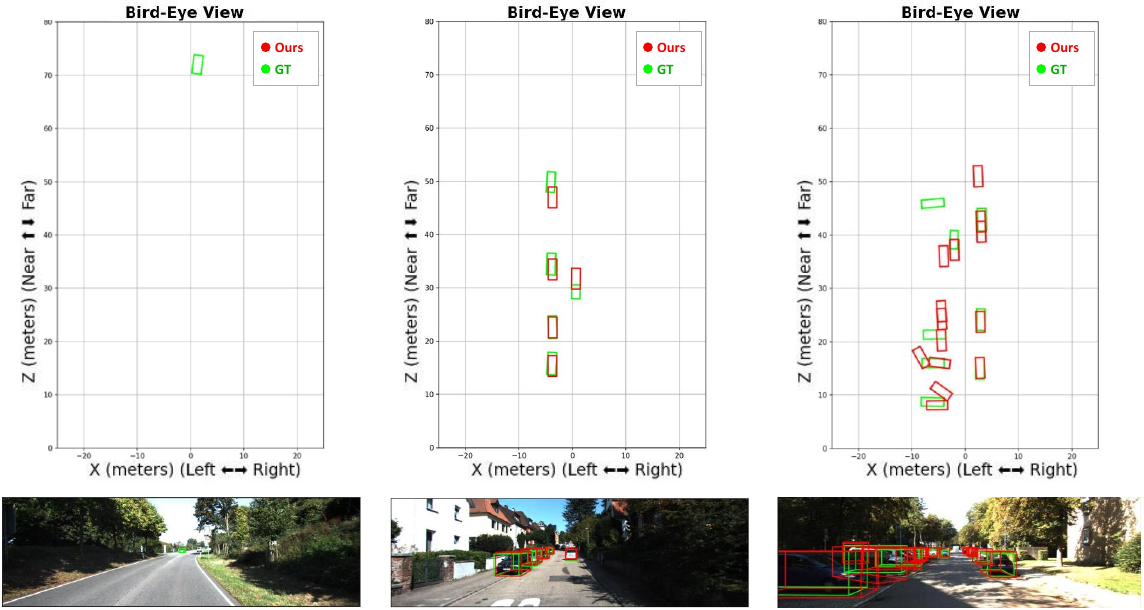}
    
    \begin{tabular}{>{\centering\arraybackslash}p{0.3\linewidth}
                >{\centering\arraybackslash}p{0.3\linewidth}
                >{\centering\arraybackslash}p{0.3\linewidth}}
    \scriptsize (1) & \scriptsize (2) & \scriptsize (3) \\
    \end{tabular}
    \caption{Failure modes analysis on KITTI~\cite{kitti}.}
    \label{fig:failure_modes_1}
\end{minipage}\hfill
\begin{minipage}[bt]{0.28\linewidth}
    \centering
    \includegraphics[width=\linewidth]{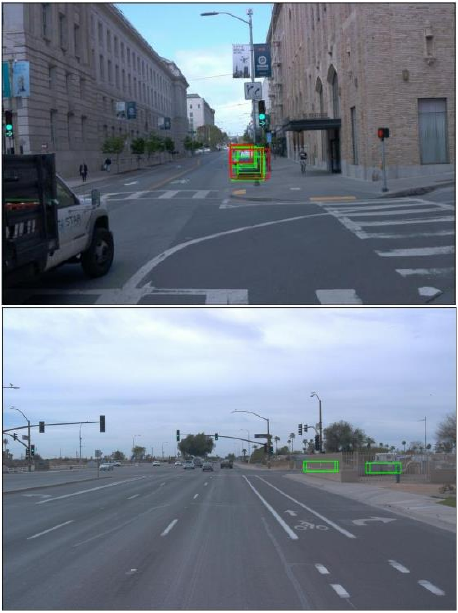}
    \caption{Failure cases in Waymo-Open~\cite{waymo}.}
    \label{fig:failure_modes_2}
\end{minipage}
\vspace{-0.8em}
\end{figure*}

\section{Failure Mode Analysis}
\label{appendix:failure}
While the quantitative and qualitative results in the main paper demonstrate the robustness of our method, it does not perform flawlessly in every scenario. This section analyzes representative failure cases, as well as situations where the proposed approach cannot theoretically provide substantial benefits. As summarized in Fig.~\ref{fig:failure_modes_1}, our failure modes fall into three major categories. 

\subsubsection{Case 1: Limited Relative Motion.}
Our approach captures and fuses shape variations through relative motion between the camera and objects. However, when the relative pose remains nearly constant--for instance, due to minimal ego-motion or objects moving steadily at a distance—the method offers limited advantages, as shown in Fig.~\ref{fig:failure_modes_1}-(1). In such cases, long-range objects provide insufficient evidence for label refinement because 2D detections remain sparse across frames, often resulting in detection failures. This behavior is evident in Fig.~\ref{fig:failure_modes_2}, where the top example from Waymo~\cite{waymo} shows a case in which a subsequent right turn of the camera causes the object to move out of view, leaving the occlusion unresolved, and the bottom example depicts an object persistently hidden by occlusion, preventing reliable observations. 

\subsubsection{Case 2: Long-range Depth Degradation.}
As with other open-set methods or monocular depth estimators, depth noise grows with distance from the camera, leading to degraded accuracy in center estimation. As shown in Fig.~\ref{fig:failure_modes_1}-(2), even though our method generally produces accurate labels, the estimated center gradually shifts as objects recede. While such issues may be alleviated as monocular depth models improve, they remain a practical bottleneck in long-range scenarios. 

\begin{wrapfigure}[15]{r}{0.48\textwidth}
    \centering
    \captionsetup{font={small}, skip=8pt}
    \vspace{-3mm}
    \includegraphics[width=0.48\textwidth]{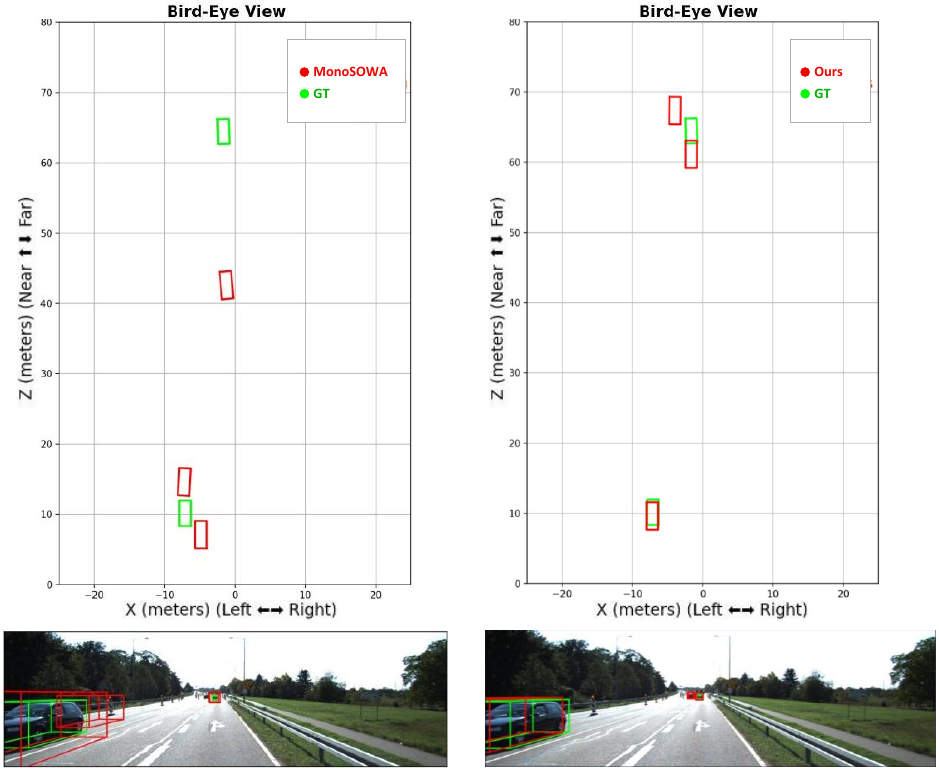}
    \begin{tabular}{>{\centering\arraybackslash}p{0.22\textwidth}>{\centering\arraybackslash}p{0.22\textwidth}}
     \scriptsize MonoSOWA\cite{skvrna2025monosowa} & \scriptsize \textbf{PLOT~(Ours})\\
    \end{tabular}
    \caption{Association failures with naive tracking. Zoom for the best view.}
    \label{fig:bev_monosowa}
\end{wrapfigure}

\subsubsection{Case 3: Edge-case Association Errors.}
Although the proposed Global Object Memory (GOM) substantially improves object association compared to conventional approaches (see Fig.~\ref{fig:bev_monosowa}), it cannot resolve all cases. Fig.~\ref{fig:failure_modes_1}-(3) illustrates such an edge case: objects located near the boundary of the tracking window with minimal observations, or objects persistently embedded in clutter (similar to Case~1), may fail to be associated correctly. This leads to inaccurate pseudo-LiDAR fusion and, consequently, erroneous 3D bounding box estimates. These cases highlight the limits of our current association strategy and suggest that integrating stronger priors on temporal continuity or leveraging multi-view geometric cues could further enhance robustness. 

\begin{figure*}[hbt] \centering
    \includegraphics[width=\textwidth]{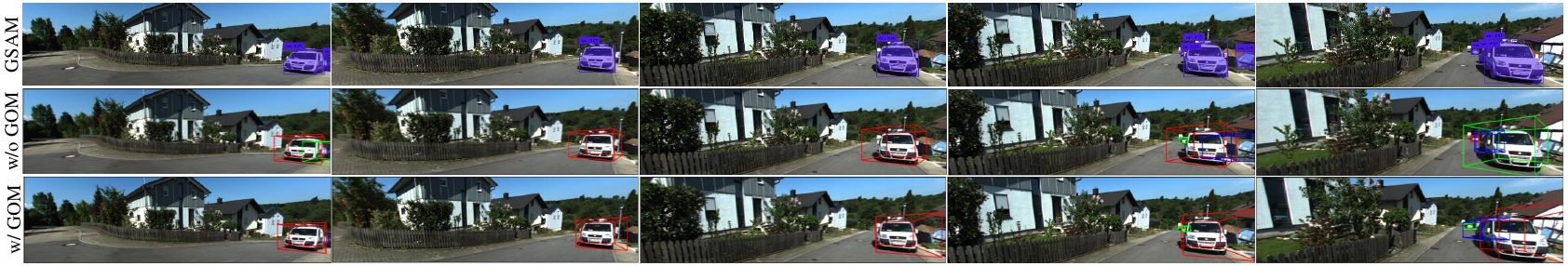}
    \caption{Qualitative comparisons with and without global object memory, on KITTI-360~\cite{kitti360}. Zoom for the best view.} 
    \label{fig:gom_ablation_qual}
\end{figure*}

\begin{figure*}[hbt] \centering
\includegraphics[width=\textwidth,height=0.35\textwidth]{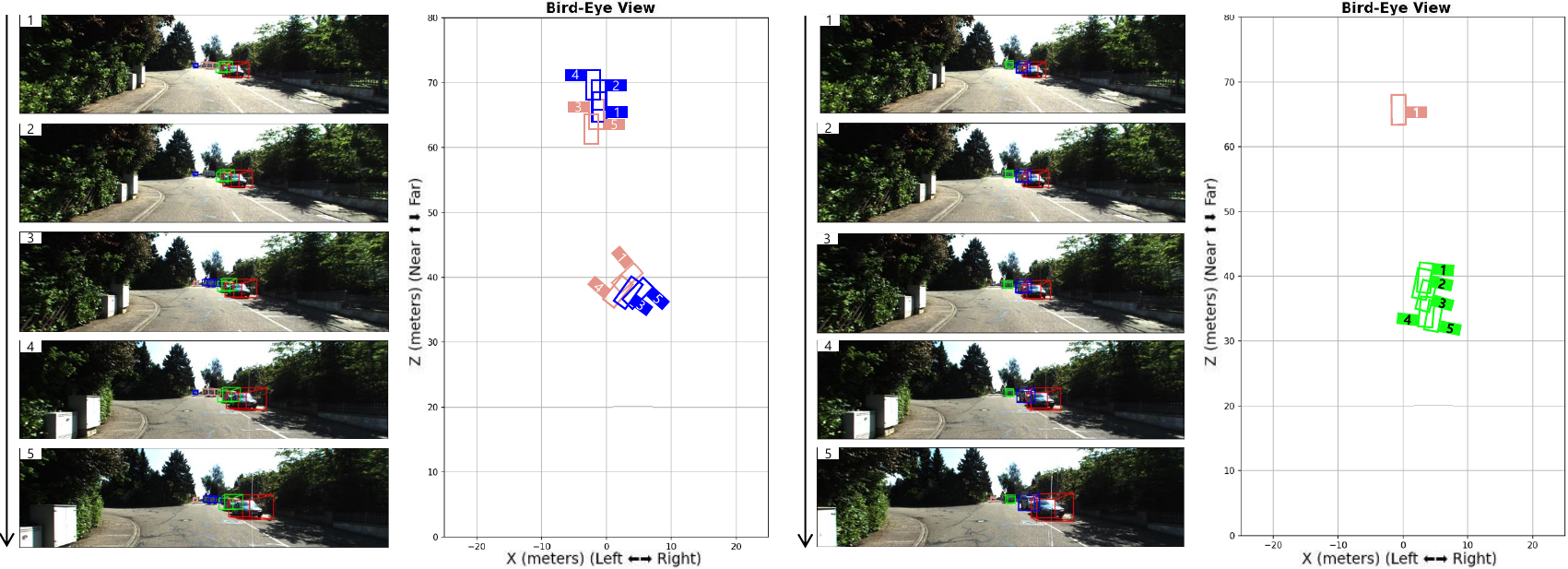}
     \caption{\textbf{Effect of accurate object association on identity consistency.} BEV visualizations with (Left Fig.) and without (Right Fig.) GOM show how ID switches are reduced when associations are corrected. Zoom for the best view.}
    \label{fig:id_switch}
\end{figure*}

\section{Additional Qualitative Results}
\label{appendix:qual}

\subsection{Necessity of Global Object Memory}
As a complement to the global object memory ablation in Tab. 5 of the main paper, we provide additional qualitative analysis in Fig.~\ref{fig:bev_monosowa} and Fig.\ref{fig:gom_ablation_qual}. Fig.~\ref{fig:bev_monosowa} shows the limitation of naive tracking-based association, while Fig.\ref{fig:gom_ablation_qual} illustrates per-frame results over five consecutive frames, showing (top) per-frame predicted masks from GSAM~\cite{ren2024grounded}, (middle) 3D bounding boxes estimated without global object memory (GOM), and (bottom) 3D bounding boxes estimated with GOM. With GOM, the predicted boxes become progressively more accurate and consistent in size and orientation as camera motion provides additional observations. In contrast, without GOM, incorrect associations give rise to erroneous boxes and ID switches. The importance of consistent association for stable shape fusion is further supported by Fig.\ref{fig:id_switch}. When shape fusion is performed based on object associations on the left figure, severe pseudo-LiDAR distortions arise, which in turn lead to errors in subsequent attribute estimation.

\begin{figure*}[tb] \centering \includegraphics[width=\textwidth,height=0.35\textwidth]{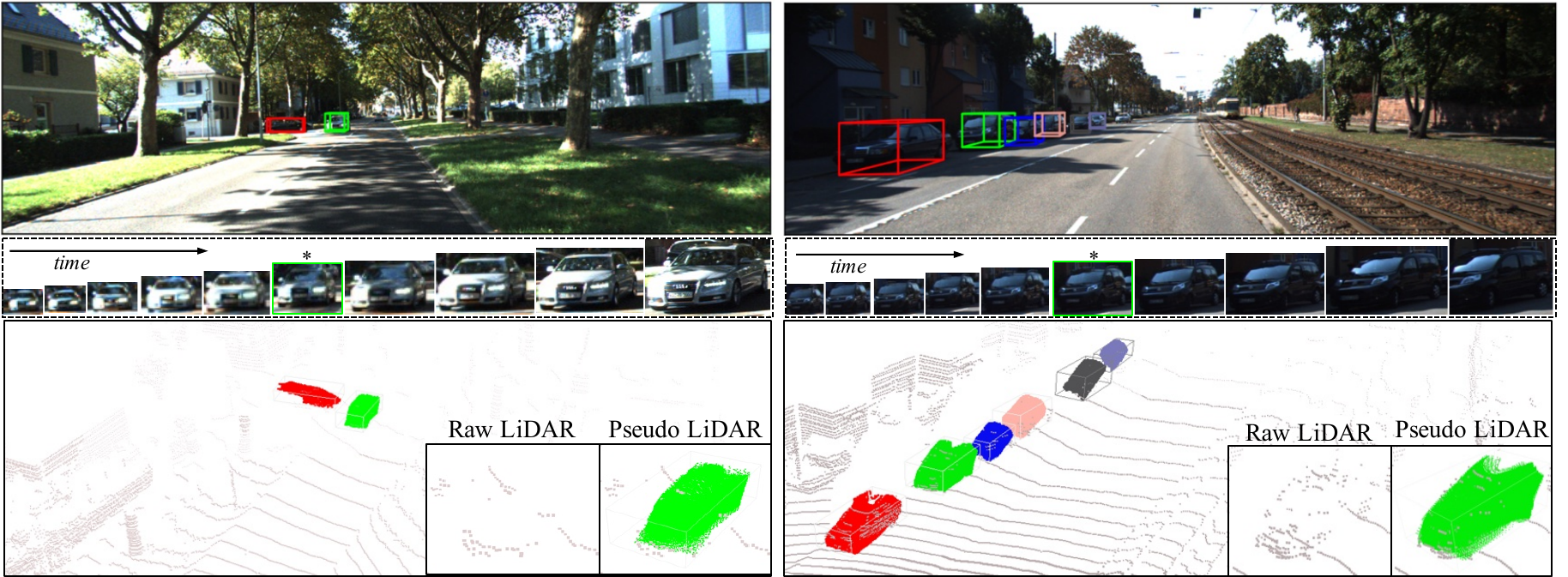}
    \caption{Completed pseudo-LiDAR obtained by trajectory-guided shape fusion.
    The middle row shows an object's multiple observations across time through ego-motion, which enables reliable orientation and dimension estimation.
    } 
    \label{fig:shape_completeness}
\end{figure*}

\subsection{Efficacy of Trajectory-Guided Shape Fusion}
Fig.~\ref{fig:shape_completeness} illustrates the efficacy of our trajectory-guided shape fusion on the KITTI~\cite{kitti} dataset. By aggregating pseudo-LiDAR observations across tracked object trajectories, our method progressively recovers more complete object geometry from multiple viewpoints. 
This improved shape completeness provides stronger geometric evidence for downstream attribute estimation, resulting in more stable orientation, dimension, and center predictions compared to partial observations from individual frames.

\subsection{Qualitative Results on KITTI}
\begin{figure*}[t] \centering
    \begin{tabular}{>{\centering\arraybackslash}p{0.25\textwidth}>{\centering\arraybackslash}p{0.25\textwidth}>{\centering\arraybackslash}p{0.24\textwidth}>{\centering\arraybackslash}p{0.23\textwidth}}
       \scriptsize Ground Truth  & \scriptsize \fade{3D-MOOD~\cite{yang20253d}} & \scriptsize MonoSOWA~\cite{skvrna2025monosowa} & \scriptsize \textbf{PLOT~(Ours})\\
    \end{tabular}
    \includegraphics[width=\textwidth]{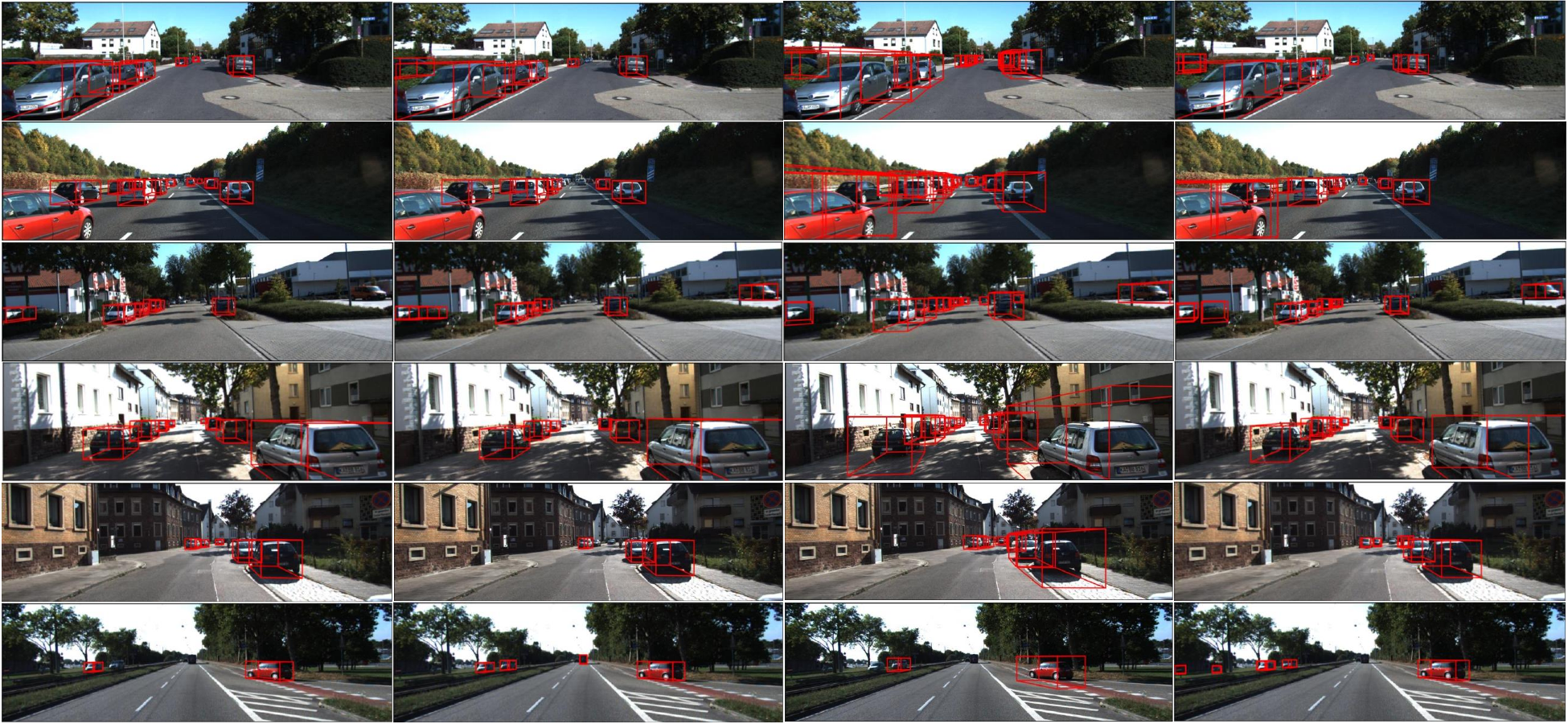}
    \caption{Qualitative comparisons on KITTI~\cite{kitti}. The supervised method- 3D-MOOD~\cite{yang20253d} is shown in gray.} 
    \label{fig:app_kitti}
\end{figure*}

Fig.~\ref{fig:app_kitti} presents qualitative comparisons on KITTI~\cite{kitti} against ground truth, a supervised open-set detector (3D-MOOD)~\cite{yang20253d}, and a pseudo-labeling method (MonoSOWA)~\cite{skvrna2025monosowa}. Notably, PLOT produces 3D boxes that are visually on par with 3D-MOOD, despite the latter being trained with full 3D supervision. In contrast, MonoSOWA frequently generates oversized boxes, reflecting the limitations of its LLM-driven~\cite{achiam2023gpt} shape priors. These results show that PLOT achieves high reliability without 3D supervision while also exposing the shortcomings of LLM-driven geometric reasoning.

\subsection{Qualitative Results on Waymo-Open}
\begin{figure*}[t] \centering
    \includegraphics[width=\textwidth]{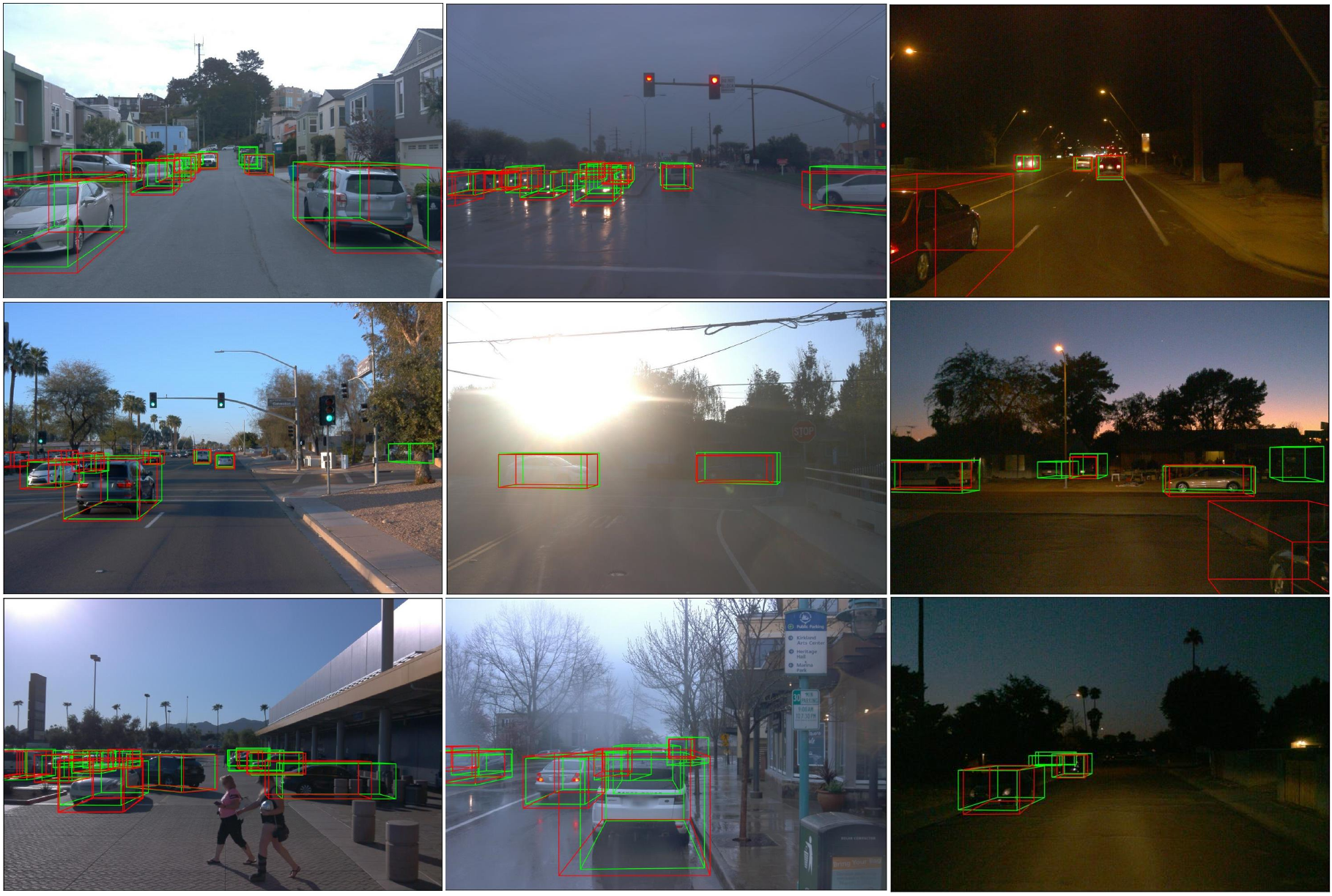}
    \caption{Qualitative results on Waymo~\cite{waymo} under challenging conditions. Left: standard driving scenes. 
Middle: scenes with challenging environmental conditions~(e.g. raining, light-saturation). Right: night scenes with limited illumination. Our trajectory-guided labeling improves temporal consistency and yields reliable pseudo-labels even in such adverse scenarios. Ground truth boxes are shown in green and predicted boxes in red.} 
    \label{fig:app_waymo}
\end{figure*}

Fig.~\ref{fig:app_waymo} presents qualitative results on Waymo-Open~\cite{waymo}, across diverse conditions: a daytime scene (left), a challenging environment with rain or light saturation (middle), and a nighttime scene (right). Ground-truth annotations are shown in green boxes, while our predicted results are shown in red. In both the daytime and challenging-condition scenes, our method produces detections that closely align with the ground truth, even under heavy clutter and adverse conditions. Remarkably, in the nighttime scene, our trajectory-guided labeling captures additional object instances that are missing in the ground-truth labels, demonstrating the robustness of our approach under low-visibility settings. These qualitative results highlight the practical potential of our method for monocular 3D object detection in real-world scenarios.

\subsection{In-the-wild Videos} 
\begin{figure*}[tb] \centering
    \begin{tabular}{>{\centering\arraybackslash}p{0.3\textwidth}>{\centering\arraybackslash}p{0.3\textwidth}>{\centering\arraybackslash}p{0.3\textwidth}}
      \scriptsize OVM3D-Det~\cite{huangtraining} & \scriptsize 3D-MOOD~\cite{skvrna2025monosowa} & \scriptsize \textbf{PLOT~(Ours})\\
    \end{tabular}
    \includegraphics[width=0.9\textwidth]{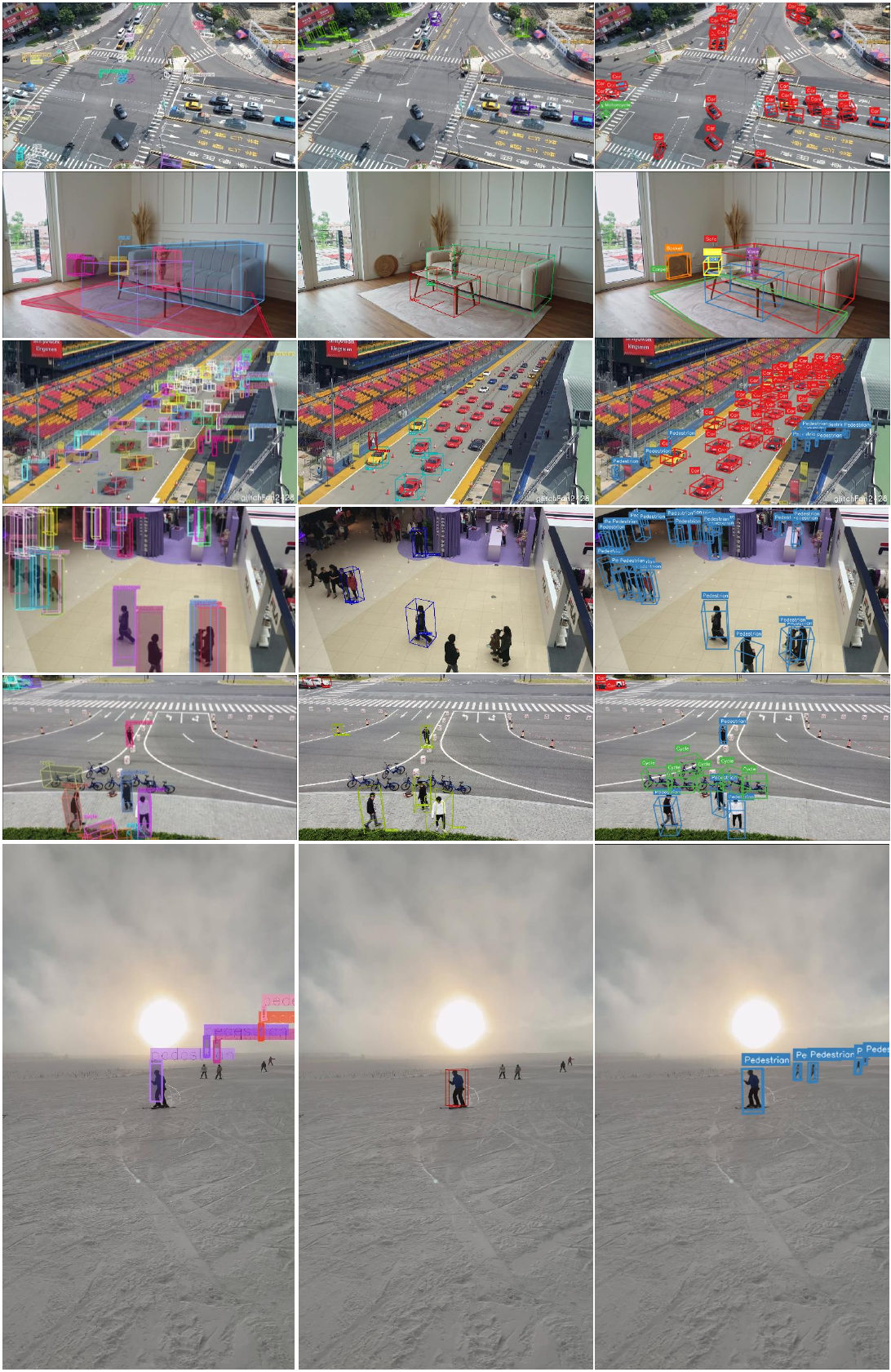}
    \caption{Additional qualitative results on in-the-wild videos. Prompts with same object classes are used for all methods.} 
    \label{fig:app_wild}
\end{figure*}

\begin{figure*}[t] \centering
\begin{tabular}{>{\centering\arraybackslash}p{0.45\textwidth}>{\centering\arraybackslash}p{0.45\textwidth}}
       \scriptsize 3D-MOOD~\cite{yang20253d} & \scriptsize \textbf{PLOT~(Ours})  \\
    \end{tabular}
    \includegraphics[width=\textwidth]{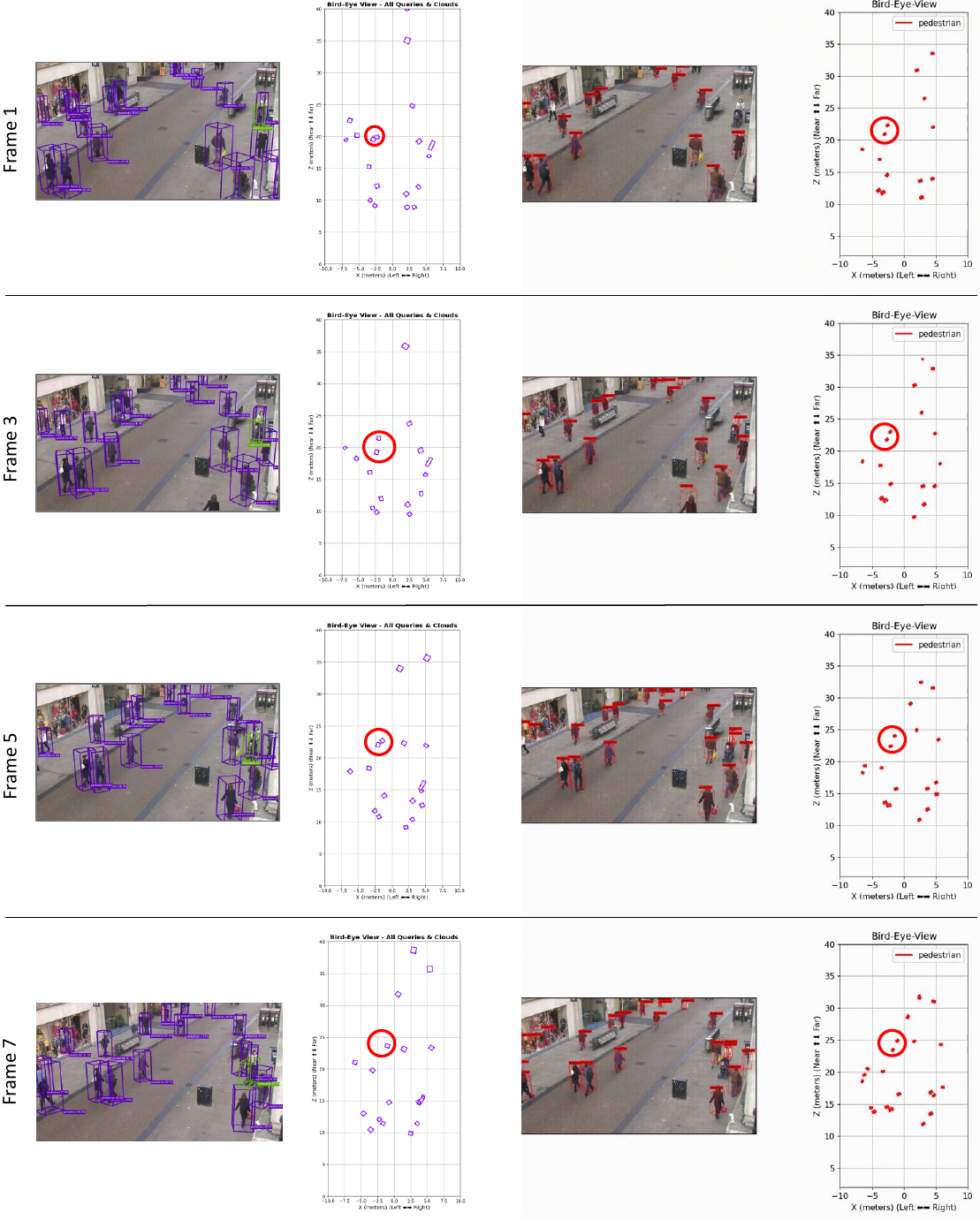}
    \vspace{-1em}
    \caption{Additional qualitative comparisons on MOT17~\cite{MOT16} sequence visualized on BEV. The red dotted circles highlight regions where our method maintains temporally consistent ground-plane geometry and stable 3D box estimates, in contrast to the spatial drift observed in 3D-MOOD~\cite{yang20253d}.
    } 
    \label{fig:app_wild_bev}
\end{figure*}

In addition to the cross-domain results shown in Fig.~1, Fig.~2, and Fig.~7 of the main paper, we compare our raw pseudo-labels with those from the open-set single-image pseudo-labeler~(OVM3D-Det)~\cite{huangtraining} and the open-set 3D object detector~(3D-MOOD)~\cite{yang20253d}, in Fig.~\ref{fig:app_wild}. OVM3D-Det, relying on single-image and shape priors, consistently produces inaccurate box sizes and center estimates, while 3D-MOOD suffers from reduced recall in novel domains. In contrast, our method consistently provides reasonable estimates across diverse conditions, including challenging cases such as phone-captured vertical videos (last row). 
To further demonstrate spatial consistency in in-the-wild videos, Fig.~\ref{fig:app_wild_bev} provides both 2D projections and BEV visualizations for a scene from MOT17~\cite{MOT16}.
From the BEV view, objects are shown to lie on a consistent ground plane with well-aligned relative positions, offering clearer evidence of the geometric coherence provided by our pipeline, whereas 3D-MOOD mainly struggles with relative spatial positioning across frames.


%% file: tab/tab6_depth_tracker.tex
\begin{tabular}{ccccc|c|c|*{2}{c}|*{2}{c}}
\toprule
\multirow{2}{*}{Alignment}
& Range
& Patch
& Temporal
& Tiny
& Depth
& Pose 
& \multicolumn{2}{c|}{$\mathrm{AP_{3D}@0.3}\uparrow$} 
& \multicolumn{2}{c}{$\mathrm{AP_{BEV}@0.3}\uparrow$} \\
& Noise & Noise & Jitter & Tracker & RMSE $\downarrow$ & ATE $\downarrow$ & Easy & Hard & Easy & Hard \\
\midrule
SE(3) & -  & - & - & \xmark & \B 3.72 & \B 0.204 & \B 20.89 & \B 14.38 & \B 27.85 & \B 20.20 \\
SE(3) & -  & - & - & \checkmark & \B 3.72 & \B 0.204 & 20.57 \red{(-0.32)} & 14.17 \red{(-0.21)} & 27.71 \red{(-0.14)} & 19.54 \red{(-0.66)} \\
\midrule
\midrule
\multirow{3}{*}{SE(3)} & 0.04   &  0.3  & \xmark & - & 3.90 \red{(+0.18)} & 0.205 & 20.30 \red{(-0.59)} & 14.00 \red{(-0.38)} & 28.80 \red{(+0.95)} & 19.73 \red{(-0.47)} \\
& 0.08   & 0.3   & \xmark & - & 4.34 \red{(+0.62)} & \B 0.204 & 19.48 \red{(-1.41)} & 13.45 \red{(-0.93)} & 26.63 \red{(-1.22)} & 19.35 \red{(-0.85)}\\
& 0.08      & 0.6   & \xmark & - & 4.38 \red{(+0.66)} & 0.206 & 19.65 \red{(-1.24)} & 13.39 \red{(-0.99)} & 26.94 \red{(-0.91)} & 19.39 \red{(-0.81)} \\
\midrule
\multirow{2}{*}{SE(3)} & \multirow{2}{*}{\xmark} & \multirow{2}{*}{\xmark} & 0.02 & - & 3.78 \red{(+0.06)} & 0.205 & 17.50 \red{(-3.39)}  & 11.96 \red{(-2.42)} & 25.84 \red{(-2.01)} & 17.43 \red{(-2.77)} \\
& & & 0.04 & - & 3.82 \red{(+0.10)} & \B 0.204 & 14.14 \red{(-6.75)} & 9.49 \red{(-4.89)} & 20.29 \red{(-7.56)} & 14.54 \red{(-5.66)} \\
\midrule
\multirow{2}{*}{Sim(3)} & \multirow{2}{*}{\xmark} & \multirow{2}{*}{\xmark} & 0.02 & - & 3.78 \red{(+0.06)} & \B 0.204 & 17.69 \red{(-3.20)}  & 13.39 \red{(-0.99)} & 25.96 \red{(-1.89)} & 19.01 \red{(-1.19)} \\
& & & 0.04 & - & 3.81 \red{(+0.09)} & \B 0.204 & 14.99 \red{(-5.90)} & 11.23 \red{(-3.15)} & 22.55 \red{(-5.30)} & 16.36 \red{(-3.84)} \\

\bottomrule
\end{tabular}

%% file: tab/atab2_orient.tex
\begin{tabular}{c|*{3}{c}|*{3}{c}}
\toprule
\multirow{2}{*}{Cam. motion} & \multicolumn{3}{c|}{ $\mathrm{AP_{3D}@IoU=0.3}$$\uparrow$} & \multicolumn{3}{c}{$\mathrm{AP_{BEV}@IoU=0.3}$$\uparrow$}  \\ 
 & Easy & Mod. & Hard & Easy & Mod. & Hard \\ 
\midrule
\xmark & 77.65 & 57.59 & 48.27 & 80.51 & 60.25 & 50.81 \\
\checkmark & \B 80.48 & \B 60.83 & \B 51.49 & \B 83.06 & \B 63.59 & \B 54.15 \\
\bottomrule
\end{tabular}

%% file: tab/atab5_intrinsics.tex
\begin{tabular}{c|*{3}{c}|*{3}{c}}
\toprule
\multirow{2}{*}{Intrinsics} & \multicolumn{3}{c|}{ $\mathrm{AP_{3D}@IoU=0.3}$$\uparrow$} & \multicolumn{3}{c}{$\mathrm{AP_{BEV}@IoU=0.3}$$\uparrow$}  \\ 
 & Easy & Mod. & Hard & Easy & Mod. & Hard \\ 
\midrule
UniDepth\cite{unidepth} & 70.44 & 51.90 & 44.12 & 76.17 & 58.66 & 50.49 \\
GT & \B 80.48 & \B 60.83 & \B 51.49 & \B 83.06 & \B 63.59 & \B 54.15 \\
\bottomrule
\end{tabular}